\PassOptionsToPackage{unicode}{hyperref}
\PassOptionsToPackage{hyphens}{url}
\documentclass[
  10pt,letterpaper,
]{article}
\usepackage{xcolor}
\usepackage[letterpaper,textwidth=345pt,textheight=550pt,
  left=134pt,top=125pt,headheight=12pt,headsep=25pt,footskip=30pt]{geometry}

\usepackage{amsmath,amssymb}
\usepackage{amsthm}
\usepackage{graphicx}
\usepackage{caption}
\everydisplay\expandafter{\the\everydisplay
  \abovedisplayshortskip=\abovedisplayskip
  \belowdisplayshortskip=\belowdisplayskip
}
\newtheorem{theorem}{Theorem}[section]
\usepackage{algorithm}
\usepackage{algpseudocode}
\usepackage{iftex}
\ifPDFTeX
  \usepackage[OT1]{fontenc} 
  \usepackage[utf8]{inputenc}
  \usepackage{textcomp} 
\else 
  \usepackage{unicode-math} 
  \defaultfontfeatures{Scale=MatchLowercase}
  \defaultfontfeatures[\rmfamily]{Ligatures=TeX,Scale=1}
\fi
\ifPDFTeX\else
\fi
\IfFileExists{upquote.sty}{\usepackage{upquote}}{}
\IfFileExists{microtype.sty}{
  \usepackage[expansion=false]{microtype}
  \UseMicrotypeSet[protrusion]{basicmath} 
}{}
\usepackage{needspace}
\usepackage{bookmark}
\IfFileExists{xurl.sty}{\usepackage{xurl}}{} 
\hypersetup{
  pdftitle={Selective Inference for Shapley-Based Feature Selection in Nonlinear Regression Models: A Loss-Reduction Approach},
  pdfauthor={Kota SHIDA, Tomohiro SHIRAISHI, Shuichi NISHINO, Ichiro TAKEUCHI},
  hidelinks,
  pdfcreator={LaTeX via pandoc}}

\title{Selective Inference for Shapley-Based Feature Selection in Nonlinear Regression Models: A Loss-Reduction Approach}
\author{Kota SHIDA, Tomohiro SHIRAISHI,\\
  Shuichi NISHINO, Ichiro TAKEUCHI}
\date{\today}

\begin{document}

\newgeometry{textwidth=345pt,textheight=620pt,left=134pt,top=55pt}
\maketitle
\thispagestyle{empty}

\begin{abstract}
Shapley-value-based approaches evaluate feature importance by aggregating contributions across different subsets of features, providing a flexible framework for capturing nonlinear relationships and feature interactions.
However, features selected based on high importance scores are not necessarily statistically significant, and valid inference must account for the data-dependent selection process.
In this paper, we propose a selective inference (SI) method for Shapley-based feature selection, focusing on the contributions of individual features to reducing regression loss.
Our approach uses nonparametric basis-function regression to quantify nonlinear feature effects and aggregates their contributions across feature subsets using Shapley weights.
By expressing the resulting importance scores as quadratic forms in the response, we characterize the feature-selection event through quadratic inequalities and derive exact conditional tests based on a truncated chi distribution.
Specifically, we consider an $L_2$-regularized radial basis function (RBF) network with fixed basis functions.
Under the standard SI setting of a fixed design and independent Gaussian noise with known variance, our method guarantees exact finite-sample control of the conditional type I error rate.
We evaluate its empirical performance through synthetic simulations and applications to benchmark datasets.

\end{abstract}
\clearpage

\restoregeometry

\section{Introduction}\label{introduction}

Evaluating the importance of individual features is a central task in data analysis, particularly when the relationships between features and the response are nonlinear or when features are correlated.
Shapley-value-based approaches provide a way to evaluate feature importance while considering different combinations of the other features \cite{fryer2021shapley}.
For example, SHAP measures how each feature contributes to a prediction from a fitted model \cite{lundberg2017unified}, whereas SAGE attributes predictive performance, measured by a loss function, to individual features \cite{covert2020understanding}.
The common idea behind these Shapley-value-based approaches is to assess a target feature across all subsets of the remaining features and aggregate its contributions using Shapley weights.
By considering multiple feature subsets rather than a single model, these approaches provide a more comprehensive assessment of feature importance, potentially capturing nonlinear relationships and feature interactions represented by the underlying models.

Even when a feature receives a high Shapley-value-based importance score from a complex nonlinear model, the score may reflect noise in the data rather than a genuine population-level effect.
It is therefore important to assess the statistical significance of selected features, especially when they inform high-stakes decisions such as medical diagnosis or autonomous driving.
For features selected by a data-dependent algorithm, the sampling distribution of a test statistic generally differs from that for features specified in advance.
Selective inference (SI) provides a principled framework for conducting valid statistical inference in such settings by conditioning on the selection event \cite{taylor2015statistical,lee2016exact,loftus2015groups}.
A key challenge in SI is to characterize the feature-selection event and derive the conditional sampling distribution of a test statistic for the target effect.
Much of the classical SI literature has focused on feature selection in linear regression, where methods such as the Lasso are used to select features and statistical inference is subsequently performed on the coefficients of the selected linear model.
In contrast, Shapley-based feature selection evaluates each feature by aggregating its contributions across different subsets of the other features, potentially using complex nonlinear models.
Extending SI to this setting is nontrivial because both the feature-selection process and the target effect depend on contributions evaluated across multiple feature subsets.

In this study, we develop an SI method for Shapley-based feature selection, focusing on a loss-based formulation inspired by SAGE.
Specifically, we consider a setting in which features are selected according to Shapley-weighted reductions in penalized regression loss.
Our goal is to assess the statistical significance of the selected features while accounting for this selection process.
Our approach is based on two key ideas.
First, we use basis-function regression to quantify nonlinear feature contributions across different subsets of features.
Second, we show that the resulting Shapley-weighted loss reductions can be expressed as quadratic forms in the response, allowing us to characterize the feature-selection event through quadratic inequalities and construct conditional tests based on a truncated chi distribution.
Together, these ideas enable statistically valid inference for features selected by aggregating nonlinear contributions across different feature subsets.

As a concrete realization of this framework, we consider an $L_2$-regularized radial basis function (RBF) network with fixed basis functions.
We formulate statistical hypotheses for the resulting Shapley-based feature effects and develop an exact conditional selective inference procedure.
Under a standard SI setting of a fixed design with independent Gaussian noise of known variance, the proposed method controls the conditional type I error rate exactly in finite samples.
We evaluate its empirical performance through synthetic simulations and applications to benchmark datasets.
For reproducibility, our implementation is available at {\urlstyle{tt}\url{https://github.com/kotashida/si-shapley-feature-selection}}.

\paragraph{Related Works.}\label{related-work}

The Shapley value, originally introduced in cooperative game theory, allocates an overall value to individual players by averaging their marginal contributions across possible coalitions \cite{shapley1953value}.
In machine learning, game-theoretic methods for explaining predictions were developed in early work such as \cite{strumbelj2010efficient}, followed by SHAP for explaining individual model predictions \cite{lundberg2017unified} and SAGE for attributing global predictive performance to features \cite{covert2020understanding}.
Unlike feature importance assessed relative to a single feature subset, these approaches aggregate contributions over different subsets, although their interpretation depends on the definition of the coalition value and the treatment of correlated features \cite{janzing2020feature}.
Their use for feature selection also raises questions about how attribution principles relate to the goals of selecting predictive variables \cite{fryer2021shapley}.
Several studies have examined the statistical reliability of Shapley-based importance measures.
Covert and Lee quantified uncertainty due to numerical estimation of Shapley values \cite{covert2021improving}; Williamson and Feng developed asymptotic inference for population-level Shapley variable importance \cite{williamson2020efficient}; and Whitehouse et al.\ studied asymptotic inference for aggregated SHAP attributions \cite{whitehouse2026statistical}.
Other approaches include randomization tests related to Shapley explanations \cite{teneggi2023shapxrt} and statistical guarantees for feature-importance rankings \cite{goldwasser2025statistical}.
Kim et al.\ proposed SHAP-guided feature selection followed by inference on linear surrogate coefficients, treating the SHAP screening set as fixed in their original selective inference formulation \cite{kim2025phitest}.
Their revised approach focuses on nominal uncertainty estimates without finite-sample selection-adjusted guarantees.
In contrast, our method explicitly accounts for Shapley-based feature selection and provides exact conditional inference for nonlinear feature effects.

Selective inference (SI) addresses data-dependent hypothesis selection by conditioning on the selection event, with foundational developments for linear-model feature selection, including the Lasso \cite{taylor2015statistical,lee2016exact,tibshirani2016exact}.
Its scope has expanded beyond feature selection to clustering \cite{gao2024selective}, changepoint detection \cite{duy2020computing,jewell2022testing,yamada2025change}, and anomaly detection \cite{tsukurimichi2022conditional,niihori2025quantifying}.
More complex selection procedures have also been considered, including boosting \cite{rugamer2020inference}, regression trees \cite{neufeld2022treevalues}, kernel-based feature selection \cite{yamada2018post}, and neural-network-based methods \cite{duy2022quantifying,miwa2023salient,shiraishi2024statistical,shiraishi2026statistical}.
Parametric-programming and homotopy-based approaches make such inference computationally tractable by tracking selection events as the response varies along a one-dimensional path \cite{le2021parametric,sugiyama2021more,duy2022quantifying}.
For tests based on linear contrasts, SI commonly uses a truncated normal distribution, while tests of group effects with quadratic selection events can use a truncated chi distribution \cite{lee2016exact,loftus2015groups}.
In our setting, Shapley-weighted reductions in penalized regression loss are quadratic forms in the response, and both the feature ranking and tuning-parameter selection can be represented by quadratic inequalities.
We exploit this structure along a one-dimensional response path to obtain exact conditional tests based on a truncated chi distribution, accounting for the Shapley-based selection procedure.

\section{Shapley-Based Feature Selection via Loss Reduction}
\label{problem-setup}\label{feature-selection}

Consider a regression dataset $(X,\boldsymbol y)$ with $n$ observations and $m$ candidate features indexed by $[m]:=\{1,\ldots,m\}$, where $X=[\boldsymbol x_1,\ldots,\boldsymbol x_m]\in\mathbb R^{n\times m}$ and $\boldsymbol y\in\mathbb R^n$.
Our goal is to select features according to their contributions to reducing regression loss.
The contribution of a feature generally depends on which other features are included in the regression model.
For example, correlated features may provide overlapping information, while nonlinear interactions can affect the contributions of individual features.
Motivated by loss-based Shapley approaches such as SAGE \cite{covert2020understanding} and work on Shapley-based feature selection \cite{fryer2021shapley}, we evaluate each target feature across different subsets of the remaining features and aggregate its contributions using Shapley weights, rather than relying on a single regression model.
In this section, we first formulate feature contributions using general nonlinear regression models, define Shapley-weighted importance scores, and describe a feature-selection procedure based on these scores.

\subsection{Nonlinear Regression Models for Feature Subsets}
\label{selection-models}

Following cooperative game theory, we call a subset of the candidate features a \emph{coalition}.
For a target feature $j\in[m]$, we consider coalitions $S\subseteq[m]\setminus\{j\}$ and refer to the features in $S$ as \emph{nuisance features}, because they serve as a reference for assessing the additional contribution of $j$.
In particular, the empty coalition $S=\varnothing$ represents a baseline model without any candidate features, whereas the full coalition $S=[m]\setminus\{j\}$ contains all candidate features except the target feature $j$.

For each coalition $S$, we consider two regression models: one fitted using only the features in $S$, and another that additionally incorporates the target feature $j$.
Our formulation is applicable to a variety of nonlinear regression models without being restricted to a specific model class.
Let $\lambda$ denote the hyperparameters of the regression model.
For a fixed $\lambda$, let $L_\lambda(S;\boldsymbol y)$ denote the loss obtained by fitting a regression model using the features in coalition $S$, and let $L_\lambda^+(j\mid S;\boldsymbol y)$ denote the corresponding loss when the target feature $j$ is additionally included.

The contribution of feature $j$ relative to coalition $S$ is measured by the loss reduction
\begin{equation}
 \label{eq:general-coalition-improvement}
  \Delta_{j\mid S}^\lambda(\boldsymbol y) = L_\lambda(S;\boldsymbol y) - L_\lambda^+(j\mid S;\boldsymbol y).
\end{equation}
A large positive value indicates that incorporating $j$ substantially reduces the regression loss in the presence of $S$.
Throughout this study, we assume that the loss reduction is nonnegative, i.e., $\Delta_{j\mid S}^\lambda(\boldsymbol y)\ge 0$, for every target feature $j$ and coalition $S$.
This is a natural assumption, as adding a feature should not increase the loss when the augmented model can ignore the additional feature if it provides no benefit.

\subsection{Shapley-Weighted Loss Reduction}
\label{shapley-selection-scores}

A contribution based on one particular coalition can give an incomplete assessment of feature importance.
For example, a feature may be highly useful when other correlated features are absent but add little when those features are already available.
Shapley weighting accounts for this variation by combining feature contributions across all subsets of the other features.

For a nonempty feature universe $U\subseteq[m]$ and a target feature $j\in U$, consider every coalition $S\subseteq U\setminus\{j\}$.
The corresponding Shapley weight is given by
\begin{equation}
\label{eq:shapley-weights}
 w_{j,U}(S) = \frac{|S|!(|U|-|S|-1)!}{|U|!} = \frac{1}{|U|\binom{|U|-1}{|S|}},
\end{equation}
where $|\cdot|$ denotes the cardinality of a set.
This weight represents the probability that the set of features preceding feature \(j\) in a uniformly random ordering of \(U\) is exactly \(S\).
Indeed, there are $|S|!$ ways to order the features in $S$ before $j$ and $(|U|-|S|-1)!$ ways to order the remaining features after $j$, out of $|U|!$ possible orderings.
In particular, each possible coalition size receives the same total weight of $1/|U|$, distributed uniformly among coalitions of that size.

We define the Shapley-weighted loss-reduction score by aggregating the contributions in \eqref{eq:general-coalition-improvement} across coalitions:
\begin{equation}
 \label{eq:shapley-loss-score}
  T_{j,U}^\lambda(\boldsymbol y) = \sum_{S\subseteq U\setminus\{j\}} w_{j,U}(S)\Delta_{j\mid S}^\lambda(\boldsymbol y).
\end{equation}
This score measures the overall contribution of feature $j$ to reducing regression loss, accounting for different subsets of the other features.

\subsection{Feature Selection}

We select the top-scoring features based on the Shapley-weighted loss-reduction scores defined in \eqref{eq:shapley-loss-score}.
Specifically, we consider the full feature universe $U=[m]$, so that each candidate feature is evaluated across all subsets of the remaining features.
The hyperparameters of the regression models are selected by cross-validation (CV), and we denote the selected configuration by $\widehat\lambda(\boldsymbol y)$.

We rank the candidate features in decreasing order of $T_{j,[m]}^{\widehat\lambda(\boldsymbol y)}(\boldsymbol y)$, breaking ties in favor of the smaller feature index.
For a prespecified number $k\in[m]$, let $M(\boldsymbol y)\subseteq[m]$ denote the set of the $k$ highest-scoring features.
Our goal is to assess the statistical significance of the selected features by computing $p$-values that account for both the feature-selection process and the cross-validation-based hyperparameter selection (see Section~\ref{selective-inference}).

\subsection{Regularized Nonparametric Regression with Fixed Basis Functions}
\label{rbf-selection}

The Shapley-based feature-selection procedure described above can be applied to a variety of nonlinear regression models.
As a concrete demonstration, we consider penalized nonparametric regression with fixed basis functions.
For each coalition $S$, we fit a regression model using basis functions of the features in $S$ and compare it with an augmented model that additionally includes basis functions of the target feature $j$.
The basis functions for $S$ are kept unchanged in this comparison, so the two models are nested.
Since both models minimize the same penalized objective and the augmented model allows the additional coefficients to be zero, the nonnegativity assumption is satisfied.
With fixed basis functions and $L_2$-regularized least squares, the loss reductions can be expressed as quadratic forms in the response.
Consequently, the feature-selection event can be characterized through quadratic inequalities, enabling the exact selective inference developed in Section~\ref{selective-inference}.
We also use this fixed-basis regression formulation to define the inferential targets in Section~\ref{selection-and-hypotheses}.

The regularization parameter is selected by $F$-fold cross-validation of the full-feature regression model.
For a prespecified grid $\Lambda=\{\lambda_1,\ldots,\lambda_L\}$ with $0<\lambda_1<\cdots<\lambda_L$, let $\operatorname{CV}_\ell(\boldsymbol y)$ denote the cross-validation error for $\lambda_\ell$.
We select
\begin{equation}
\label{eq:penalty-choice}
\begin{aligned}
\widehat\ell(\boldsymbol y)
&=\underset{\ell\in\{1,\ldots,L\}}{\operatorname{argmin}}\;
\operatorname{CV}_\ell(\boldsymbol y),\\
\widehat\lambda(\boldsymbol y)
&=\lambda_{\widehat\ell(\boldsymbol y)},
\end{aligned}
\end{equation}
where ties favor the smaller penalty.
The selected hyperparameter is used for all coalition models, and its selection is accounted for in the selective inference procedure.
For the basis functions, we use Gaussian radial basis functions (RBFs) in our implementation.
The regression model and loss functions are specified in Section~\ref{coalition-models}.

\newtheorem{lemma}[theorem]{Lemma}

\section{Statistical Hypotheses for Selected Features}
\label{selection-and-hypotheses}

Section~\ref{feature-selection} describes how a set of features $M(\boldsymbol y)$ is selected using Shapley-weighted loss-reduction scores.
Throughout, we distinguish the random response vector $\boldsymbol Y$, a generic realization $\boldsymbol y$, and the observed realization $\boldsymbol y_{\rm obs}$.
Accordingly, $M(\boldsymbol Y)$ is a random selected feature set, $M(\boldsymbol y)$ is the set selected for a given response $\boldsymbol y$, and $M_{\rm obs}=M(\boldsymbol y_{\rm obs})$ is the observed selected feature set.
We consider a feature $j\in M_{\rm obs}$ chosen for testing.
Our goal is to assess the statistical significance of the population-level effect of this selected feature.

Once a subset of features has been selected, it is natural to evaluate the importance of each feature in the context of the other selected features.
We therefore consider a Shapley-weighted effect that aggregates the contributions of $j$ across all coalitions $S\subseteq M_{\rm obs}\setminus\{j\}$.
This allows us to assess how much the selected feature contributes to reducing regression loss across different combinations of the other selected features, accounting for potentially overlapping information and nonlinear relationships among them.

To quantify this effect, we use $L_2$-regularized nonparametric regression with fixed basis functions.
This choice of inferential model is conceptually separate from the algorithm used for feature selection, although the present study uses the same fixed-basis formulation for both purposes.
We first define the population-level effect and corresponding statistical hypotheses, and then derive a test statistic with a tractable null distribution when the inferential target is fixed.
The data-dependent selection process is accounted for in Section~\ref{selective-inference}.

\subsection{Statistical Model}
\label{statistical-model}

We treat the design matrix $X$ as fixed and assume
\begin{equation}
\label{eq:model}
\boldsymbol Y=\boldsymbol\mu(X)+\boldsymbol\varepsilon,
\qquad
\boldsymbol\varepsilon\sim\mathcal N(\boldsymbol0,\sigma^2 I_n),
\qquad
\sigma>0\ \text{known}.
\end{equation}
Here $\boldsymbol\mu(X)\in\mathbb R^n$ is the unknown mean response, which we abbreviate as $\boldsymbol\mu=\boldsymbol\mu(X)$.
This is known as the \emph{saturated model} setting in selective inference \cite{fithian2014optimal}, where the mean vector $\boldsymbol\mu$ is unrestricted.
This setting is flexible because it does not require the regression models used for feature selection or inference to be correctly specified.
In particular, the basis functions introduced below define the effects of interest, but the true mean need not lie in their span.

We assume that the basis functions and cross-validation folds are determined before observing the response and are treated as fixed throughout the analysis.
Let $\lambda_{\rm obs}=\widehat\lambda(\boldsymbol y_{\rm obs})$ denote the regularization parameter selected by cross-validation.
The observed selected feature set $M_{\rm obs}$, target feature $j$, and regularization parameter $\lambda_{\rm obs}$ specify the inferential target.
Although these quantities are treated as fixed when defining the target, their data dependence is accounted for by conditioning on the selection event in Section~\ref{selective-inference}.

\subsection{Population Feature Effects and Statistical Hypotheses}
\label{statistical-hypotheses}

\paragraph{Nonparametric regression for a target feature and its nuisance coalition.}
\label{coalition-models}
For a target feature $j$ and a nuisance coalition $S$ with $j \notin S$, we represent the effects of the features in $S$ jointly using multivariate basis functions, and the additional effect of $j$ using univariate basis functions.
The multivariate representation allows nonlinear relationships and interactions among nuisance features, while the univariate component provides a common way to measure the added contribution of $j$ in each coalition.
By aggregating these comparisons over coalitions, we obtain a Shapley-weighted assessment that does not depend on a single choice of nuisance features.

For each nonempty subset $S\subseteq[m]$, let $p_S$ denote the number of basis functions used for the features in $S$, and let $\{\phi_{S,q}:q\in[p_S]\}$ denote these basis functions, where $[p_S]:=\{1,\ldots,p_S\}$.
Write $X_{i,S}$ for the feature vector of observation $i$ restricted to $S$, and define the centered basis matrix
\begin{equation}
 \label{eq:basis}
 (G_S)_{iq} =\phi_{S,q}(X_{i,S}) -\frac1n\sum_{a=1}^n\phi_{S,q}(X_{a,S}), \qquad G_S\in\mathbb R^{n\times p_S}.
\end{equation}
For $S=\varnothing$, let $G_\varnothing$ have zero columns, and for a single target feature $j$, write $G_j=G_{\{j\}}$ and $p_j=p_{\{j\}}$.
Thus, $G_S$ may contain multivariate functions of the nuisance features, whereas $G_j$ contains univariate functions of the target feature $j$.
The basis functions are constructed from $X$ according to a prespecified rule and remain fixed as the response varies.
In our numerical implementation, we use Gaussian radial basis functions (RBFs), although the theoretical results do not depend on this particular choice.

For a fixed penalty $\lambda>0$, the nuisance-only model has minimized penalized training loss
\begin{equation}
 \label{eq:coalition-objective}
  L_\lambda(S;\boldsymbol y) =\min_{\beta_0,\boldsymbol\gamma} \left\{\|\boldsymbol y-\beta_0\mathbf1_n-G_S\boldsymbol\gamma\|_2^2 +n\lambda\|\boldsymbol\gamma\|_2^2 \right\},
\end{equation}
where $\mathbf1_n\in\mathbb R^n$ denotes the vector of ones, $\beta_0\in\mathbb R$, and $\boldsymbol\gamma\in\mathbb R^{p_S}$ (with the terms involving $\boldsymbol\gamma$ omitted when $S=\varnothing$).
We measure the additional contribution of $j$ using the augmented model, which retains the same nuisance basis $G_S$ and adds $G_j$:
\begin{equation}
\label{eq:joint-coalition-loss}
\begin{aligned}
L_\lambda^+(j\mid S;\boldsymbol y) =\min_{\beta_0,\boldsymbol\gamma,\boldsymbol\beta_j}
\bigl\{
\|\boldsymbol y-\beta_0\mathbf1_n -G_S\boldsymbol\gamma-G_j\boldsymbol\beta_j\|_2^2
+n\lambda(\|\boldsymbol\gamma\|_2^2 +\|\boldsymbol\beta_j\|_2^2)
\bigr\},
\end{aligned}
\end{equation}
where $\boldsymbol\beta_j\in\mathbb R^{p_j}$.
These are the concrete losses corresponding to the general definition in Section~\ref{selection-models}.
Because the nuisance-only model is recovered by setting $\boldsymbol\beta_j=\boldsymbol0$, the loss reduction $\Delta_{j\mid S}^\lambda(\boldsymbol y)$ in \eqref{eq:general-coalition-improvement} is nonnegative.
Notice that this is the reduction in the minimized \emph{penalized training objective}, not necessarily the change in out-of-sample prediction error.

\paragraph{Population effects and hypotheses.}

For the selected target feature $j\in M_{\rm obs}$, we assess its contribution relative to the other selected features.
Specifically, we consider all nuisance coalitions $S\subseteq M_{\rm obs}\setminus\{j\}$ and aggregate the corresponding loss reductions using Shapley weights.
This provides a measure of the contribution of $j$ within the selected feature set, accounting for different combinations of the other selected features.
Note that feature selection itself uses all candidate features, as described in Section~\ref{feature-selection}.

For the observed regularization parameter $\lambda_{\rm obs}$, we define the population feature effect by evaluating the Shapley-weighted loss-reduction score in \eqref{eq:shapley-loss-score} at the noise-free mean response:
\begin{equation}
\label{eq:population-effect}
T_{j,M_{\rm obs}}^{\lambda_{\rm obs}}(\boldsymbol\mu)
=
\sum_{S\subseteq M_{\rm obs}\setminus\{j\}}
w_{j,M_{\rm obs}}(S)
\Delta_{j\mid S}^{\lambda_{\rm obs}}(\boldsymbol\mu).
\end{equation}
We consider the following hypotheses:
\begin{equation}
\label{eq:hypotheses}
H_{0,j,M_{\rm obs}}:
T_{j,M_{\rm obs}}^{\lambda_{\rm obs}}(\boldsymbol\mu)=0
\quad\text{versus}\quad
H_{1,j,M_{\rm obs}}:
T_{j,M_{\rm obs}}^{\lambda_{\rm obs}}(\boldsymbol\mu)>0.
\end{equation}
Since all coalition-level loss reductions are nonnegative and all Shapley weights are positive, the null hypothesis holds if and only if $\Delta_{j\mid S}^{\lambda_{\rm obs}}(\boldsymbol\mu)=0$ for every $S\subseteq M_{\rm obs}\setminus\{j\}$.
Thus, the null states that the target feature provides no reduction in penalized loss for the mean response across any coalition of the other selected features.
In contrast, an observed loss reduction may be positive merely because of response noise.
We next use a quadratic representation of the Shapley-weighted effect to construct a test statistic with a tractable null distribution; the derivation is provided in Appendix~\ref{app:proofs-section3}.

\subsection{Test Statistic and Its Fixed-Target Distribution}
\label{test-space}

We now consider how to test the zero-effect hypothesis in \eqref{eq:hypotheses}.
A natural statistic is the Shapley-weighted loss-reduction score $T_{j,M_{\rm obs}}^{\lambda_{\rm obs}}(\boldsymbol Y)$, which measures the contribution of the selected feature $j$ across all coalitions of the other selected features.
As established in Appendix~\ref{app:proofs-section3}, this score can be expressed as a quadratic form in the response:
\[
T_{j,M_{\rm obs}}^{\lambda_{\rm obs}}(\boldsymbol Y) = \boldsymbol Y^\top A_{j,M_{\rm obs}}^{\lambda_{\rm obs}}\boldsymbol Y,
\]
where $A_{j,M_{\rm obs}}^{\lambda_{\rm obs}}$ is a symmetric positive semidefinite matrix whose explicit expression is given in \eqref{eq:shapley}.
This quadratic representation allows us to characterize the zero-effect hypothesis and construct a test statistic with a simple null distribution.

For the observed selected feature set $M_{\rm obs}$, target feature $j$, and regularization parameter $\lambda_{\rm obs}$, define the test space and its orthogonal projector by
\begin{equation}
\label{eq:test-space}
\begin{aligned}
W_j
&=\operatorname{range}(A_{j,M_{\rm obs}}^{\lambda_{\rm obs}}),\\
P_{W_j}
&=\text{orthogonal projector onto }W_j,\\
d_j
&=\dim(W_j).
\end{aligned}
\end{equation}
For simplicity, we suppress the dependence of $W_j$ on $M_{\rm obs}$ and $\lambda_{\rm obs}$.
Since the effect matrix is positive semidefinite, the zero-effect hypothesis is equivalent to a projection-based null hypothesis:
\begin{equation}
\label{eq:projection-null}
H_{0,j,M_{\rm obs}}:
\quad
T_{j,M_{\rm obs}}^{\lambda_{\rm obs}}(\boldsymbol\mu)=0
\quad\Longleftrightarrow\quad
P_{W_j}\boldsymbol\mu=\boldsymbol0.
\end{equation}
This motivates the projection-based test statistic
\begin{equation}
\label{eq:projection-statistic}
Q_j(\boldsymbol Y)
=
\frac{\|P_{W_j}\boldsymbol Y\|_2^2}{\sigma^2}.
\end{equation}
The following theorem characterizes the null distributions when the inferential target is specified independently of the response.

\begin{theorem}[Fixed-target null distributions]
\label{thm:fixed-target-chi-square}

Fix the design matrix $X$, the basis functions, a nonempty feature set $M\subseteq[m]$, a target feature $j\in M$, and a regularization parameter $\lambda>0$ independently of the response $\boldsymbol Y$.
Let $A=A_{j,M}^\lambda$, $W=\operatorname{range}(A)$, $P_W$ be the orthogonal projector onto $W$, and $d=\dim(W)$.
Under the Gaussian model \eqref{eq:model} and the null hypothesis $H_0:T_{j,M}^\lambda(\boldsymbol\mu)=0$, the Shapley-weighted loss-reduction statistic has the distribution
\begin{equation}
\label{eq:weighted-chi-square}
\frac{T_{j,M}^\lambda(\boldsymbol Y)}{\sigma^2}
\overset{d}{=}
\sum_{r=1}^{d}a_r Z_r^2,
\qquad
Z_r\overset{\mathrm{iid}}{\sim}\mathcal N(0,1),
\end{equation}
where $a_1,\ldots,a_d$ are the positive eigenvalues of $A$.
Furthermore, if $d\ge1$, the projection-based statistic satisfies
\begin{equation}
\label{eq:fixed-target-chi-square}
Q_{j,M}^\lambda(\boldsymbol Y)
:=
\frac{\|P_W\boldsymbol Y\|_2^2}{\sigma^2}
\sim\chi_d^2.
\end{equation}
For $d=0$, both statistics are identically zero.
\end{theorem}

The complete proof of Theorem~\ref{thm:fixed-target-chi-square}, together with the supporting technical results, is provided in Appendix~\ref{app:proofs-section3}.
The key observation is that the Shapley-weighted loss reduction can be expressed as a positive semidefinite quadratic form in the response.
Under the zero-effect null hypothesis, the mean response is orthogonal to the range of the corresponding matrix, so the standardized Gaussian response projected onto this space has independent standard normal coordinates.
The loss-reduction statistic is a weighted sum of their squares, whereas the projection-based statistic is their unweighted sum, yielding the respective null distributions.
These distributions apply when the inferential target is fixed; the data-dependent feature selection and hyperparameter selection are accounted for in Section~\ref{selective-inference}.

\section{Selective Inference}
\label{selective-inference}

In modern data analysis, statistical hypotheses are often formulated through data-dependent procedures, such as feature selection and pattern discovery using complex algorithms.
Since the hypotheses themselves depend on the observed data, their statistical significance cannot generally be assessed using conventional inferential procedures that assume prespecified hypotheses.
Selective inference (SI) provides a principled framework for valid statistical inference on such data-driven hypotheses by conditioning on the selection event.
By accounting for the selection process, SI enables statistically valid assessment of hypotheses discovered through data analysis.
In our setting, this requires conditioning on the selected feature set $M(\boldsymbol Y)$ and appropriately accounting for any additional data-dependent choice of a target feature $j\in M(\boldsymbol Y)$ when testing the population-level effect defined in Section~\ref{selection-and-hypotheses}.

The general principle of SI is not restricted to a particular feature-selection algorithm.
In principle, selection procedures based on complex models, such as random forests and neural networks, can also be accommodated if the corresponding selection event can be appropriately characterized and the required conditional distribution can be evaluated.
However, for such algorithms, explicit characterization of the selection event is often computationally difficult, and conditional sampling methods may instead be needed.

In this study, we focus on a tractable setting in which Shapley-based feature selection is performed using $L_2$-regularized nonparametric regression with fixed basis functions, as in the inferential model of Section~\ref{selection-and-hypotheses}.
In this setting, both the Shapley-weighted loss-reduction scores and the cross-validation criteria can be expressed as quadratic functions of the response.
Consequently, the selection event can be characterized by quadratic inequalities, allowing exact selective inference through one-dimensional interval calculations.
We first define the selection event and conditioning scheme, then derive a one-dimensional representation of the conditional distribution, and finally establish the selective $p$-value and its statistical validity.

\subsection{Selection Event and Conditioning}
\label{selection-event}

Recall that $M_{\rm obs}=M(\boldsymbol y_{\rm obs})$ denotes the observed selected feature set and $\lambda_{\rm obs}=\widehat\lambda(\boldsymbol y_{\rm obs})$ denotes the regularization parameter selected by cross-validation.
For a target feature $j\in M_{\rm obs}$, the null hypothesis in \eqref{eq:hypotheses} is defined in terms of $M_{\rm obs}$ and $\lambda_{\rm obs}$, both of which depend on the observed response.
Therefore, the fixed-target null distribution in Theorem~\ref{thm:fixed-target-chi-square} cannot be used directly without accounting for these data-dependent choices.

We consider the selection event
\begin{equation}
\label{eq:selection-event}
\mathcal E_{\rm obs}
=
\left\{
M(\boldsymbol Y)=M_{\rm obs},
\quad
\widehat\lambda(\boldsymbol Y)=\lambda_{\rm obs}
\right\}.
\end{equation}
Conditioning on this event fixes the selected feature set and regularization parameter, and hence the population effect and test space associated with the target feature $j$.
In particular, because $j\in M_{\rm obs}$, the event $M(\boldsymbol Y)=M_{\rm obs}$ also accounts for the fact that $j$ was selected.
Here we assume that the target feature is prespecified or chosen using a rule that depends only on the selected feature set, without any additional response-dependent selection.
If $j$ is chosen using an additional data-dependent rule, such as selecting the highest-scoring feature within $M_{\rm obs}$, the outcome of that rule must also be included in the conditioning event.

We now consider the specific feature-selection procedure based on fixed-basis nonparametric regression.
As described in Section~\ref{feature-selection}, the regularization parameter is selected by cross-validation, and the $k$ features with the largest Shapley-weighted loss-reduction scores are selected from all $m$ candidate features.
For a fixed regularization parameter, each coalition-level loss reduction is a quadratic function of the response, and the corresponding Shapley-weighted score is therefore also quadratic.
Likewise, because the basis functions and cross-validation folds are fixed, the fitted responses in each fold depend linearly on $\boldsymbol y$, making the cross-validation criteria quadratic functions of $\boldsymbol y$.

Consequently, the event $\widehat\lambda(\boldsymbol Y)=\lambda_{\rm obs}$ can be expressed through quadratic inequalities comparing the cross-validation criteria of candidate regularization parameters.
Conditional on this regularization parameter, the event $M(\boldsymbol Y)=M_{\rm obs}$ can be expressed through pairwise comparisons of Shapley-weighted scores between selected and unselected features:
\begin{equation}
\label{eq:ranking-inequalities}
T_{a,[m]}^{\lambda_{\rm obs}}(\boldsymbol Y)
\ge
T_{b,[m]}^{\lambda_{\rm obs}}(\boldsymbol Y),
\qquad
a\in M_{\rm obs},\quad b\notin M_{\rm obs},
\end{equation}
where ties are resolved using a prespecified rule.
Here the scores in \eqref{eq:ranking-inequalities} are used for feature selection over all candidate features, whereas the inferential target in \eqref{eq:hypotheses} is defined relative to the selected feature set $M_{\rm obs}$.
Thus, the selection event $\mathcal E_{\rm obs}$ is characterized by a finite collection of quadratic inequalities in $\boldsymbol Y$.
This structure allows the conditional distribution required for SI to be evaluated through the one-dimensional reduction described next.

\subsection{Test Statistic and One-Dimensional Reduction}
\label{test-statistic-and-conditioning}

Recall from Section~\ref{test-space} that the zero-effect hypothesis for the selected feature $j$ can be expressed as $H_{0,j,M_{\rm obs}}:P_{W_j}\boldsymbol\mu=\boldsymbol0$, where $W_j=\operatorname{range}(A_{j,M_{\rm obs}}^{\lambda_{\rm obs}})$.
Let $P=P_{W_j}$ denote the orthogonal projector onto this test space, and let $d_j=\operatorname{rank}(P)$.
We first consider the nondegenerate case $d_j\ge1$.

The projection-based test statistic introduced in \eqref{eq:projection-statistic} can be written as
\[
Q_j(\boldsymbol Y)
=
\frac{\|P\boldsymbol Y\|_2^2}{\sigma^2}
=
R_j(\boldsymbol Y)^2,
\]
where the radial statistic $R_j(\boldsymbol Y)$, the orthogonal component $\boldsymbol Z_j$, and the direction $\boldsymbol D_j$ are defined by
\begin{equation}
\label{eq:radial-statistic}
\begin{aligned}
R_j(\boldsymbol Y)
&=\frac{\|P\boldsymbol Y\|_2}{\sigma},\\
\boldsymbol Z_j
&=(I_n-P)\boldsymbol Y,\\
\boldsymbol D_j
&=\frac{P\boldsymbol Y}{\|P\boldsymbol Y\|_2}.
\end{aligned}
\end{equation}
Under the null hypothesis $P\boldsymbol\mu=\boldsymbol0$, and when the test space is fixed, $R_j(\boldsymbol Y)$ follows the chi distribution $\chi_{d_j}$ and is independent of $\boldsymbol Z_j$ and $\boldsymbol D_j$.
The key idea is to condition on the observed orthogonal component and direction, leaving only the radial statistic as a random quantity.

Specifically, when $P\boldsymbol y_{\rm obs}\ne\boldsymbol0$ (which holds almost surely under \eqref{eq:model} when $d_j\ge1$), define
\[
\boldsymbol z_j=(I_n-P)\boldsymbol y_{\rm obs},
\qquad
\boldsymbol u_j=
\frac{P\boldsymbol y_{\rm obs}}{\|P\boldsymbol y_{\rm obs}\|_2},
\qquad
r_{\rm obs}=
\frac{\|P\boldsymbol y_{\rm obs}\|_2}{\sigma};
\]
if $P\boldsymbol y_{\rm obs}=\boldsymbol0$, we set $p_j^{\rm SI}=1$.
Conditioning on $\boldsymbol Z_j=\boldsymbol z_j$ and $\boldsymbol D_j=\boldsymbol u_j$ restricts the possible response vectors to the one-dimensional ray
\begin{equation}
\label{eq:response-line}
\boldsymbol y_j(r)
=
\boldsymbol z_j+\sigma r\boldsymbol u_j,
\qquad r\ge0.
\end{equation}
The observed response corresponds to $r=r_{\rm obs}$.

On this ray, the original Shapley-weighted loss-reduction score satisfies
\begin{equation}
\label{eq:effect-on-ray}
T_{j,M_{\rm obs}}^{\lambda_{\rm obs}}(\boldsymbol y_j(r))
=
\sigma^2r^2
\boldsymbol u_j^\top
A_{j,M_{\rm obs}}^{\lambda_{\rm obs}}
\boldsymbol u_j.
\end{equation}
Since $\boldsymbol u_j\in W_j$ and $d_j\ge1$, the quadratic coefficient in \eqref{eq:effect-on-ray} is strictly positive.
Thus, on the conditioned ray, the original Shapley-weighted loss-reduction score and the radial statistic have the same ordering, and their upper-tail tests yield the same selective $p$-value.

The selection event further restricts the admissible values of $r$.
Define the selection region
\begin{equation}
\label{eq:selection-region}
\mathcal Z
=
\left\{
r\ge0:
M(\boldsymbol y_j(r))=M_{\rm obs},
\quad
\widehat\lambda(\boldsymbol y_j(r))=\lambda_{\rm obs}
\right\}.
\end{equation}
By construction, $r_{\rm obs}\in\mathcal Z$.
Since the selection event is characterized by quadratic inequalities in the response, substituting \eqref{eq:response-line} transforms these inequalities into quadratic inequalities in the scalar variable $r$.
Therefore, $\mathcal Z$ can be represented as a union of intervals whose endpoints are obtained from the roots of these quadratic inequalities.
This reduction makes it possible to compute the exact conditional distribution of the radial statistic without sampling from the original $n$-dimensional response space.

\subsection{Selective \texorpdfstring{\(p\)}{p}-Value and Validity}
\label{selective-p-value-and-validity}

We now derive the conditional distribution of the radial statistic and the corresponding selective $p$-value.
Let $f_{\chi_{d_j}}$ denote the probability density function of the chi distribution with $d_j$ degrees of freedom, and let $\mathbf1\{\cdot\}$ denote the indicator function.
The following theorem establishes the conditional validity of the resulting test.

\begin{theorem}[Conditional distribution and validity]
\label{thm:selective-validity}

Assume the Gaussian model \eqref{eq:model}, with fixed design, prespecified basis functions and cross-validation folds, and known noise variance $\sigma^2$.
Consider the observed selection event $\mathcal E_{\rm obs}$ in \eqref{eq:selection-event} and a target feature $j\in M_{\rm obs}$ whose choice requires no additional response-dependent conditioning.
Let $P=P_{W_j}$ and assume $d_j=\operatorname{rank}(P)\ge1$.
Under the null hypothesis $H_{0,j,M_{\rm obs}}:P\boldsymbol\mu=\boldsymbol0$, the conditional density of $R_j(\boldsymbol Y)$ given $\mathcal E_{\rm obs}$, $\boldsymbol Z_j=\boldsymbol z_j$, and $\boldsymbol D_j=\boldsymbol u_j$ is
\begin{equation}
\label{eq:conditional-density}
f(r\mid\mathcal E_{\rm obs},\boldsymbol Z_j=\boldsymbol z_j,\boldsymbol D_j=\boldsymbol u_j;H_{0,j,M_{\rm obs}})
=
\frac{
f_{\chi_{d_j}}(r)\mathbf1\{r\in\mathcal Z\}
}{
\displaystyle\int_{\mathcal Z}f_{\chi_{d_j}}(t)\,dt
},
\end{equation}
for almost every nondegenerate conditioning value for which the denominator is positive.
Define the selective $p$-value as the upper-tail probability
\begin{equation}
\label{eq:selective-pvalue}
p_j^{\rm SI}(\boldsymbol Y)
=
\frac{
\displaystyle\int_{\mathcal Z\cap[R_j(\boldsymbol Y),\infty)}
f_{\chi_{d_j}}(t)\,dt
}{
\displaystyle\int_{\mathcal Z}
f_{\chi_{d_j}}(t)\,dt
},
\end{equation}
with observed value $p_j^{\rm SI}=p_j^{\rm SI}(\boldsymbol y_{\rm obs})$ obtained by replacing $R_j(\boldsymbol Y)$ with $r_{\rm obs}$.
For any $\alpha\in(0,1)$ and any mean vector $\boldsymbol\mu$ satisfying the null hypothesis, this $p$-value satisfies
\begin{equation}
\label{eq:conditional-validity}
\mathbb P_{\boldsymbol\mu}
\bigl(
p_j^{\rm SI}(\boldsymbol Y)\le\alpha
\mid\,
\mathcal E_{\rm obs},\,
\boldsymbol Z_j=\boldsymbol z_j,\,
\boldsymbol D_j=\boldsymbol u_j
\bigr)
\le\alpha.
\end{equation}
Thus, the test controls the conditional type I error rate at the nominal significance level.
\end{theorem}

The result follows from the independence of the radial statistic and the conditioned orthogonal component and direction under the null hypothesis.
The selection event restricts the radius to $\mathcal Z$, so its conditional distribution is the chi distribution truncated to this region.
The selective $p$-value is then obtained from the upper tail of this truncated distribution, yielding the conditional validity guarantee.
A complete proof is provided in Appendix~\ref{appendix:proof-selective-validity}.

At a prespecified significance level $\alpha\in(0,1)$, we reject the null hypothesis when $p_j^{\rm SI}\le\alpha$.
For the degenerate case $d_j=0$, the test statistic is identically zero, and we set $p_j^{\rm SI}=1$.
The validity guarantee is exact in finite samples under the stated Gaussian model and exact computation of the selection region.
If the selection region is approximated numerically or estimated using sampling, the same exact guarantee does not automatically follow.
Furthermore, the guarantee applies to an individual selected feature whose selection has been appropriately accounted for; controlling the error rate for multiple reported discoveries requires additional considerations.

\section{Computation of Selective \texorpdfstring{\(p\)}{p}-Values}
\label{selection-region-computation}

In this section, we present a computational procedure for evaluating the selective $p$-value derived in Section~\ref{selective-inference}.
Our approach builds on the parametric programming framework for selective inference \cite{duy2021parametric,duy2022generalized}, which characterizes the selection event by varying the response along a one-dimensional path.
In our setting, the Shapley-based feature-selection event and the cross-validation event can both be expressed as quadratic inequalities in the response.
By restricting these inequalities to the one-dimensional path in \eqref{eq:response-line}, we can efficiently identify the truncation region $\mathcal Z$ and evaluate the selective $p$-value through integration over the resulting intervals.

\subsection{Quadratic Representation of the Joint Selection Event}
\label{quadratic-selection-event}

To evaluate the selective $p$-value in \eqref{eq:selective-pvalue}, we identify the radii $r\ge0$ along the ray $\boldsymbol y_j(r)$ in \eqref{eq:response-line} that preserve both the observed selected feature set $M_{\rm obs}$ and the cross-validated regularization parameter $\lambda_{\rm obs}$.
As described in Section~\ref{selection-event}, the selection event can be represented by quadratic inequalities corresponding to feature-score comparisons and cross-validation comparisons.

\paragraph{Feature-selection comparisons.}

Keeping the selected feature set equal to $M_{\rm obs}$ requires every selected feature to have a Shapley-weighted score at least as large as that of every unselected feature.
Thus, for every $a\in M_{\rm obs}$ and $b\in[m]\setminus M_{\rm obs}$, we require
\begin{equation}
\label{eq:topk-comparison}
\boldsymbol y_j(r)^\top
\left(
A_{a,[m]}^{\lambda_{\rm obs}}
-
A_{b,[m]}^{\lambda_{\rm obs}}
\right)
\boldsymbol y_j(r)
\ge0.
\end{equation}
Here $A_{a,[m]}^{\lambda_{\rm obs}}$ denotes the quadratic-form matrix of the Shapley-weighted score used for selecting feature $a$ from all candidate features, as defined by \eqref{eq:shapley} with the feature universe $[m]$.
These selection-score matrices differ in general from the matrix $A_{j,M_{\rm obs}}^{\lambda_{\rm obs}}$ defining the population effect and test space for the selected feature $j$.
The matrices used in \eqref{eq:topk-comparison} must be constructed using the same fixed basis functions and numerical conventions as the actual feature-selection procedure.

\paragraph{Cross-validation comparisons.}

For each candidate regularization parameter $\lambda\in\Lambda$, let $E_\lambda\in\mathbb R^{n\times n}$ denote the matrix mapping the response $\boldsymbol y$ to its out-of-fold residuals.
Because the basis functions and cross-validation folds are fixed independently of the response, the fitted validation responses and their residuals depend linearly on $\boldsymbol y$.
In the RBF implementation, the basis centers, widths, and centering constants are computed from the training feature vectors within each fold and then applied to the validation feature vectors.
The full-feature regression model uses the objective in \eqref{eq:coalition-objective} with $S=[m]$, with the ridge penalty scaled by the number of training observations in each fold.

The pooled cross-validation mean squared error is therefore a quadratic function of $\boldsymbol y$:
\begin{equation}
\label{eq:cv}
\operatorname{CV}_\ell(\boldsymbol y)
=
\boldsymbol y^\top C_{\lambda_\ell}\boldsymbol y,
\qquad
C_\lambda
=
\frac1n E_\lambda^\top E_\lambda
\succeq0.
\end{equation}
Here each validation observation receives weight $1/n$, including when the cross-validation folds have different sizes.
Keeping the cross-validated regularization parameter equal to $\lambda_{\rm obs}$ requires
\begin{equation}
\label{eq:cv-comparison}
\boldsymbol y_j(r)^\top
\left(
C_\lambda-C_{\lambda_{\rm obs}}
\right)
\boldsymbol y_j(r)
\ge0,
\qquad
\lambda\in\Lambda\setminus\{\lambda_{\rm obs}\}.
\end{equation}
If the regularization parameter is specified in advance rather than selected by cross-validation, these comparisons are omitted.
When cross-validation is used, both the feature-selection comparisons in \eqref{eq:topk-comparison} and the cross-validation comparisons in \eqref{eq:cv-comparison} are required.
If equality occurs, the prespecified tie-breaking rules determine whether the observed selection is preserved.

\paragraph{Reduction to quadratic inequalities.}

Let $D\in\mathbb R^{n\times n}$ denote any symmetric matrix difference appearing in \eqref{eq:topk-comparison} or \eqref{eq:cv-comparison}.
Substituting $\boldsymbol y_j(r)=\boldsymbol z_j+\sigma r\boldsymbol u_j$ from \eqref{eq:response-line}, each comparison reduces to a quadratic inequality in $r$:
\begin{equation}
\label{eq:quadratic-comparison}
\sigma^2
(\boldsymbol u_j^\top D\boldsymbol u_j)r^2
+
2\sigma
(\boldsymbol u_j^\top D\boldsymbol z_j)r
+
\boldsymbol z_j^\top D\boldsymbol z_j
\ge0.
\end{equation}
Thus, the selection region $\mathcal Z$ in \eqref{eq:selection-region} can be computed by solving a finite collection of one-dimensional quadratic inequalities.

\subsection{Computation of the Truncation Intervals}
\label{computing-the-selection-region}

To construct $\mathcal Z$, we collect and sort all positive real roots of the quadratic polynomials in \eqref{eq:quadratic-comparison}.
These roots partition $[0,\infty)$ into intervals on which the sign of every nonzero comparison polynomial is constant.
We evaluate the comparisons at one interior point of each interval to determine whether both the observed selected feature set and the regularization parameter are preserved.
Intervals satisfying all comparisons, with the prescribed tie-breaking rules applied where necessary, are retained and adjacent admissible intervals are merged to obtain $\mathcal Z$.
Since the radial statistic has a continuous distribution, inclusion or exclusion of individual interval endpoints does not affect the selective $p$-value.

When the regularization parameter is selected from $L$ candidates and exactly $k$ of the $m$ features are selected, the number of comparisons is
\begin{equation}
\label{eq:number-of-comparisons}
N_{\mathrm{cmp}}
=
(L-1)+k(m-k).
\end{equation}
Each nonzero comparison polynomial has degree at most two and therefore contributes at most two positive real roots.
Consequently, the roots partition $[0,\infty)$ into at most $2N_{\mathrm{cmp}}+1$ open intervals.
If a comparison polynomial is identically zero, the corresponding features or regularization parameters remain tied along the entire ray, and the prescribed tie-breaking rule determines whether the observed selection is preserved.
This bound concerns only the number of intervals examined after the quadratic-form matrices have been constructed.
The exhaustive computation of Shapley-weighted loss-reduction scores still involves $2^{m-1}$ nuisance coalitions per candidate feature.

\subsection{Evaluation and Overall Algorithm}
\label{numerical-evaluation}

Once the truncation region $\mathcal Z$ has been obtained, the selective $p$-value in \eqref{eq:selective-pvalue} can be evaluated by integrating the chi density over the admissible intervals.
Let $F_d(t)=\mathbb P(\chi_d\le t)$ denote the cumulative distribution function of the chi distribution with $d$ degrees of freedom.
For an interval $[a,b]$, its probability under this distribution is
\[
\int_a^b f_{\chi_d}(t)\,dt
=
F_d(b)-F_d(a).
\]
We sum these probabilities over all intervals in $\mathcal Z$ to obtain the denominator of \eqref{eq:selective-pvalue}, and over their intersections with $[r_{\rm obs},\infty)$ to obtain the numerator.
Algorithm~\ref{alg:selective-inference} summarizes the overall procedure for testing the population-level effect of a selected feature.
Floating-point safeguards, stable probability evaluation, and further implementation details are provided in Appendix~\ref{appendix:numerical-implementation}.

\begin{algorithm}[!htbp]
\small
\caption{Selective inference after Shapley-based feature selection}
\label{alg:selective-inference}
\begin{algorithmic}[1]
\Require Data $(X,\boldsymbol y_{\rm obs})$; known noise scale $\sigma>0$; selection inputs specified in Section~\ref{feature-selection}; a prespecified rule for choosing the tested feature that depends only on the selected feature set.
\State Apply the feature-selection procedure described in Section~\ref{feature-selection} to $(X,\boldsymbol y_{\rm obs})$.
\State Set $M_{\rm obs}=M(\boldsymbol y_{\rm obs})$ and $\lambda_{\rm obs}=\widehat\lambda(\boldsymbol y_{\rm obs})$.
\State Choose a target feature $j\in M_{\rm obs}$ using the prespecified rule that depends only on $M_{\rm obs}$.
\State Construct the loss-reduction matrix $A_{j,M_{\rm obs}}^{\lambda_{\rm obs}}$ using \eqref{eq:shapley}.
\State Set $W_j=\operatorname{range}(A_{j,M_{\rm obs}}^{\lambda_{\rm obs}})$, $P=P_{W_j}$, and $d_j=\operatorname{rank}(P)$.
\If{$d_j=0$ or $P\boldsymbol y_{\rm obs}=\boldsymbol0$}
    \State \Return $1$.
\EndIf
\State Compute $r_{\rm obs}$, $\boldsymbol z_j$, and $\boldsymbol u_j$ as in Section~\ref{test-statistic-and-conditioning}.
\State Construct the quadratic comparisons \eqref{eq:topk-comparison} and \eqref{eq:cv-comparison} using the same score matrices and tie-breaking rules as in the selection procedure.
\State Compute the truncation region $\mathcal Z$ by solving the inequalities in \eqref{eq:quadratic-comparison}.
\State Evaluate the selective $p$-value $p_j^{\rm SI}$ using \eqref{eq:selective-pvalue} and the interval probabilities.
\Ensure Selective $p$-value $p_j^{\rm SI}$ for the selected feature $j$.
\end{algorithmic}
\end{algorithm}

\section{Numerical Experiments}\label{numerical-experiments}
\captionsetup[figure]{font={small,stretch=1.1},skip=5pt}
\captionsetup[table]{font={small,stretch=1.1}}
\setlength{\textfloatsep}{10pt plus 2pt minus 2pt}
\setlength{\floatsep}{8pt plus 2pt minus 2pt}
\setlength{\intextsep}{8pt plus 2pt minus 2pt}
\renewcommand{\topfraction}{0.95}
\renewcommand{\bottomfraction}{0.9}
\renewcommand{\textfraction}{0.05}
\renewcommand{\floatpagefraction}{0.7}
\setcounter{topnumber}{4}
\setcounter{totalnumber}{6}
\makeatletter
\setlength{\@fptop}{0pt}
\setlength{\@fpsep}{10pt plus 1fil}
\newif\if@floatplacedhere
\let\@orig@addtocurcol\@addtocurcol
\def\@addtocurcol{%
  \let\@prev@midlist\@midlist
  \@orig@addtocurcol
  \ifx\@midlist\@prev@midlist
    \global\@floatplacedherefalse
  \else
    \global\@floatplacedheretrue
  \fi
}
\newcommand{\@noindentafterherefloat}{%
  \if@floatplacedhere
    \global\@floatplacedherefalse
    \everypar{{\setbox\z@\lastbox}\everypar{}}%
  \fi
}
\newcommand{\@noindentafterdisplay}{%
  \@ifnextchar\par{\everypar{{\setbox\z@\lastbox}\everypar{}}}{}%
}
\AddToHook{env/figure/before}{\global\@floatplacedherefalse}
\AddToHook{env/table/before}{\global\@floatplacedherefalse}
\AddToHook{env/figure/after}{\@noindentafterherefloat}
\AddToHook{env/table/after}{\@noindentafterherefloat}
\AddToHook{env/equation/after}{\@noindentafterdisplay}
\AddToHook{env/equation*/after}{\@noindentafterdisplay}
\AddToHook{env/align/after}{\@noindentafterdisplay}
\AddToHook{env/align*/after}{\@noindentafterdisplay}
\AddToHook{env/gather/after}{\@noindentafterdisplay}
\AddToHook{env/gather*/after}{\@noindentafterdisplay}
\makeatother

We evaluate the proposed method on synthetic and real datasets in terms of type I error rate and power, testing the population-level feature effect within the selected feature set in \eqref{eq:hypotheses}.
Unless stated otherwise, we set \(\alpha=0.05\), \(k=2\), and \(p_S=\lceil 0.1n\rceil\) Gaussian RBFs per coalition, and select the regularization parameter by five-fold cross-validation over the \(L=17\) grid values \(\Lambda=\{10^{-10},10^{-9},\ldots,10^6\}\).
Given \(M(\boldsymbol Y)\), we draw one tested feature \(J\in M(\boldsymbol Y)\) uniformly at random.
For synthetic data, we set \(m=10\), generate independent rows of \(X\) from \(\mathcal N(\boldsymbol0,\Gamma)\) with \(\Gamma\in\mathbb R^{m\times m}\), \(\Gamma_{ab}=\rho^{|a-b|}\) (using \(0^0:=1\)), and \(\rho\in\{0,0.5\}\), and use noise \(\boldsymbol\varepsilon\sim\mathcal N(\boldsymbol0,I_n)\) with known \(\sigma=1\), generating 10,000 datasets per setting across \(n\in\{20,40,60\}\).
We compare the proposed selective test in \eqref{eq:selective-pvalue}, denoted Proposed, against two baselines: Naive, which uses the untruncated chi test without accounting for selection, and Bonferroni, which multiplies the naive \(p\)-value by \(Lk\binom{m}{k}\) and truncates the result at one.
Further setup, baseline, and computational resource details are given in Appendices~\ref{appendix:experimental-setup}, \ref{appendix:comparison-methods}, and~\ref{appendix:computational-resources}.

\subsection{Type I Error Rate}\label{experiments-type-i}

\begin{figure}[!htbp]
\centering
\includegraphics[width=\linewidth]{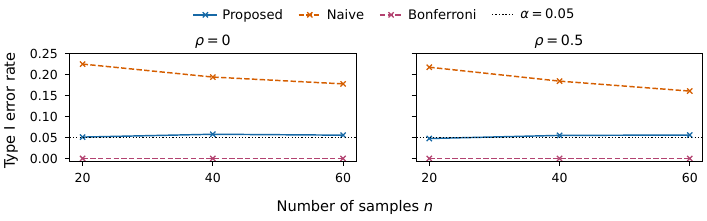}
\caption{Type I error rate conditional on \(J=1\) as the sample size \(n\) varies.
Proposed remains near the nominal level and Bonferroni is conservative across all settings, whereas Naive is anti-conservative.}\label{fig:null-main}
\end{figure}

Under the global null \(\boldsymbol\mu=\boldsymbol0\), we evaluate the rejection rate conditional on \(J=1\) across \(n\in\{20,40,60\}\) and \(\rho\in\{0,0.5\}\) in Figure~\ref{fig:null-main}.
Proposed stays near the nominal level \(\alpha=0.05\), while Naive is anti-conservative and Bonferroni is overly conservative, matching Theorem~\ref{thm:selective-validity}.
Additional null results, including sensitivity to the number of RBF basis functions (Figure~\ref{fig:sec6-basis-sensitivity}) and results stratified by the selected penalty, are provided in Appendix~\ref{appendix:additional-null}.

\subsection{Power}\label{experiments-power}

\begin{figure}[!htbp]
\centering
\includegraphics[width=\linewidth]{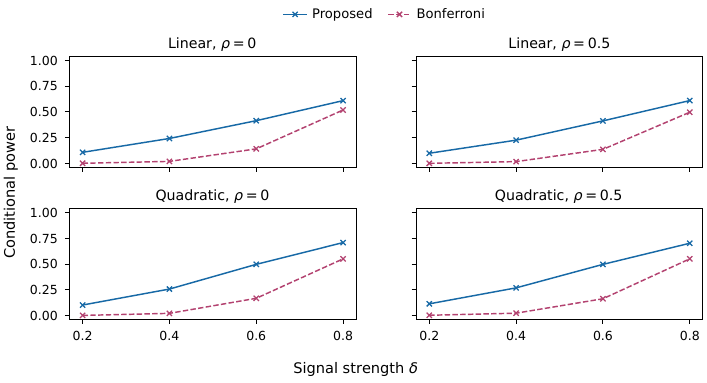}
\caption{Conditional power given \(J=1\) at \(n=40\) as the signal strength \(\delta\) varies.
Proposed achieves higher observed power than Bonferroni across all displayed settings.}\label{fig:power-selected}
\end{figure}

We generate responses from \(\boldsymbol Y=\delta\widetilde{\boldsymbol f}(\boldsymbol x_1)+\boldsymbol\varepsilon\) with \(\delta\in\{0.2,0.4,0.6,0.8\}\), where \(\widetilde{\boldsymbol f}(\boldsymbol x_1)\) applies \(f\) elementwise to \(\boldsymbol x_1\) and centers and scales the result to unit root mean square, and evaluate power conditional on \(J=1\).
In the main text, we consider the linear (\(f(t)=t\)) and quadratic (\(f(t)=t^2-1\)) signals at \(n=40\).
Because Naive is anti-conservative under the null in Figure~\ref{fig:null-main}, Figure~\ref{fig:power-selected} compares only Proposed and Bonferroni.
Proposed gains power as \(\delta\) increases and achieves higher observed power than Bonferroni across all displayed settings.
Results for the mixed and sine signals and for \(n\in\{20,60\}\) are reported in Appendix~\ref{appendix:additional-power}, with their functional forms given in Appendix~\ref{appendix:experimental-setup}.

\subsection{Real Data Experiments}\label{experiments-real-data}

We compare Proposed and Bonferroni on eight UCI regression datasets (Data1--Data8; details in Appendix~\ref{appendix:real-data-details}) across \(n\in\{20,40,60\}\), using 10,000 sub-sampled datasets per setting and standardized outcomes with \(\sigma=1\).
Under the assumption that tested selected features are relevant, we interpret rejection rates as empirical power.
Table~\ref{tab:real-main} shows that the rejection rate of Proposed increases with \(n\) on all eight datasets, achieving higher rates than Bonferroni on all eight datasets at \(n=20\) and \(n=40\) and on six of the eight datasets at \(n=60\) (with Bonferroni slightly higher on Data3 and Data5 at \(n=60\)).

\begin{table}[!htbp]\centering\small
\caption{Empirical rejection rates on eight real datasets across sample sizes \(n\).
Each cell reports Proposed/Bonferroni, with the larger unrounded value in bold.
Proposed achieves higher observed rates across all datasets at \(n=20\) and \(n=40\), and on six of the eight datasets at \(n=60\).}\label{tab:real-main}
\vspace{5pt}\setlength{\tabcolsep}{3pt}
\begin{tabular*}{\linewidth}{@{\extracolsep{\fill}}c*{8}{r}@{}}\hline
\(n\) & Data1 & Data2 & Data3 & Data4 & Data5 & Data6 & Data7 & Data8\\\hline
20 & \textbf{.21}/.03 & \textbf{.18}/.01 & \textbf{.76}/.02 & \textbf{.77}/.08 & \textbf{.74}/.02 & \textbf{.27}/.02 & \textbf{.14}/.00 & \textbf{.11}/.00\\
40 & \textbf{.44}/.15 & \textbf{.40}/.13 & \textbf{.92}/.61 & \textbf{.96}/.71 & \textbf{.92}/.58 & \textbf{.55}/.20 & \textbf{.25}/.03 & \textbf{.18}/.01\\
60 & \textbf{.59}/.33 & \textbf{.61}/.40 & .94/\textbf{.98} & \textbf{.98}/.98 & .94/\textbf{.98} & \textbf{.69}/.58 & \textbf{.37}/.09 & \textbf{.26}/.03\\
\hline\end{tabular*}\end{table}

\section{Conclusions}\label{conclusions}
In this study, we developed a selective inference method for Shapley-based feature selection via loss reduction, extending post-selection inference beyond single-model linear settings to nonlinear feature effects aggregated across multiple subsets.
Using \(L_2\)-regularized nonparametric regression with fixed basis functions, we expressed both Shapley-weighted loss-reduction scores and cross-validation criteria as quadratic forms in the response.
This characterizes the joint feature- and hyperparameter-selection event through quadratic inequalities and yields exact finite-sample selective \(p\)-values via a truncated chi distribution, controlling the conditional type I error rate at a prespecified level.

Several directions remain for future work.
First, the exact finite-sample theory assumes a fixed design and independent Gaussian noise with known variance; extending the framework to estimated variance or non-Gaussian noise remains open.
Second, exhaustive Shapley evaluation requires all \(2^m\) coalitions, so developing scalable extensions for larger numbers of candidate features is an important challenge.

\appendix


\section{Proofs and Technical Details for Section~\ref{selection-and-hypotheses}}
\label{app:proofs-section3}

This appendix provides the linear and quadratic representations underlying the test statistic in Section~\ref{test-space}, characterizes the associated test space, and proves Theorem~\ref{thm:fixed-target-chi-square}.
Throughout, $M\subseteq[m]$ denotes a fixed nonempty feature set, $j\in M$ a fixed target feature, $S\subseteq M\setminus\{j\}$ a nuisance coalition, and $\lambda>0$ a fixed regularization parameter.
The inferential target in the main text is obtained by setting $M=M_{\rm obs}$ and $\lambda=\lambda_{\rm obs}$; the fixed-target distribution theorem applies before accounting for this data-dependent selection.
The basis matrices and cross-validation folds are determined without using the response and are treated as fixed.

\subsection{Linear and Quadratic Representations}
\label{app:linear-quadratic-representations}
\label{quadratic-representation}

Define
\[
P_0=\frac{\mathbf1_n\mathbf1_n^\top}{n},\qquad
C_0=I_n-P_0,\qquad
K_S=G_SG_S^\top,
\]
and
\[
H_S^\lambda=K_S(K_S+n\lambda I_n)^{-1},\qquad
R_S^\lambda=C_0-H_S^\lambda.
\]
For $S=\varnothing$, the matrix $G_\varnothing$ has zero columns and $H_\varnothing^\lambda=0$.
Since $G_S$ is column-centered, $K_S\mathbf1_n=\boldsymbol0$ and $H_S^\lambda$ vanishes on the constant-vector subspace.
On its orthogonal complement, the eigenvalues of $H_S^\lambda$ have the form $t/(t+n\lambda)$ with $t\ge0$, so $0\preceq H_S^\lambda\preceq C_0$ and hence $R_S^\lambda\succeq0$.
Define the positive definite matrix
\[
\Omega_{j\mid S}^\lambda
=G_j^\top R_S^\lambda G_j+n\lambda I_{p_j}\succ0.
\]

\begin{lemma}[Linear representation of the target coefficients]
\label{lem:linear-target-coefficients}
The fitted coefficient vector of the target component in the augmented model \eqref{eq:joint-coalition-loss} satisfies
\begin{equation}
\label{eq:coalition-gain}
\begin{aligned}
\widehat{\boldsymbol\beta}_{j\mid S}^\lambda(\boldsymbol y)
&=B_{j\mid S}^\lambda\boldsymbol y,\\
B_{j\mid S}^\lambda
&=(\Omega_{j\mid S}^\lambda)^{-1}G_j^\top R_S^\lambda.
\end{aligned}
\end{equation}
\end{lemma}

\begin{proof}
Because the columns of $G_S$ and $G_j$ are centered, the unpenalized intercept minimizing either regression objective is the sample mean of the response.
For the nuisance-only regression, minimizing over the nuisance coefficients gives
\[
\begin{aligned}
L_\lambda(S;\boldsymbol y)
&=\min_{\boldsymbol\gamma}\left\{\|C_0\boldsymbol y-G_S\boldsymbol\gamma\|_2^2+n\lambda\|\boldsymbol\gamma\|_2^2\right\}\\
&=\boldsymbol y^\top(C_0-H_S^\lambda)\boldsymbol y
=\boldsymbol y^\top R_S^\lambda\boldsymbol y,
\end{aligned}
\]
where we used
\[
G_S(G_S^\top G_S+n\lambda I_{p_S})^{-1}G_S^\top
=K_S(K_S+n\lambda I_n)^{-1}
=H_S^\lambda
\]
for nonempty $S$, and the same identity with both sides zero for $S=\varnothing$.
For a fixed target coefficient vector $\boldsymbol\beta_j$, the minimized augmented objective equals
\[
(\boldsymbol y-G_j\boldsymbol\beta_j)^\top
R_S^\lambda(\boldsymbol y-G_j\boldsymbol\beta_j)
+n\lambda\|\boldsymbol\beta_j\|_2^2.
\]
Expanding this expression yields
\[
L_\lambda(S;\boldsymbol y)
-2\boldsymbol\beta_j^\top G_j^\top R_S^\lambda\boldsymbol y
+\boldsymbol\beta_j^\top\Omega_{j\mid S}^\lambda\boldsymbol\beta_j.
\]
Since $\Omega_{j\mid S}^\lambda\succ0$, the unique minimizer satisfies
\[
\Omega_{j\mid S}^\lambda
\widehat{\boldsymbol\beta}_{j\mid S}^\lambda(\boldsymbol y)
=G_j^\top R_S^\lambda\boldsymbol y,
\]
which proves \eqref{eq:coalition-gain}.
\end{proof}

\begin{lemma}[Loss reduction as a weighted coefficient norm]
\label{lem:weighted-loss-reduction}
The coalition-level loss reduction is
\begin{equation}
\label{eq:coalition-improvement}
\begin{aligned}
\Delta_{j\mid S}^\lambda(\boldsymbol y)
&=\bigl(\widehat{\boldsymbol\beta}_{j\mid S}^\lambda(\boldsymbol y)\bigr)^\top
\Omega_{j\mid S}^\lambda
\widehat{\boldsymbol\beta}_{j\mid S}^\lambda(\boldsymbol y)\\
&=\left\|(\Omega_{j\mid S}^\lambda)^{1/2}
\widehat{\boldsymbol\beta}_{j\mid S}^\lambda(\boldsymbol y)\right\|_2^2\\
&=\boldsymbol y^\top(B_{j\mid S}^\lambda)^\top
\Omega_{j\mid S}^\lambda B_{j\mid S}^\lambda\boldsymbol y\ge0.
\end{aligned}
\end{equation}
\end{lemma}

\begin{proof}
Let $\boldsymbol g=G_j^\top R_S^\lambda\boldsymbol y$ and $\Omega=\Omega_{j\mid S}^\lambda$.
By the preceding proof, the augmented objective after minimizing over the intercept and nuisance coefficients is
\[
L_\lambda(S;\boldsymbol y)-2\boldsymbol\beta_j^\top\boldsymbol g
+\boldsymbol\beta_j^\top\Omega\boldsymbol\beta_j.
\]
Completing the square gives
\[
\begin{aligned}
&L_\lambda(S;\boldsymbol y)-\boldsymbol g^\top\Omega^{-1}\boldsymbol g\\
&\quad +(\boldsymbol\beta_j-\Omega^{-1}\boldsymbol g)^\top
\Omega(\boldsymbol\beta_j-\Omega^{-1}\boldsymbol g).
\end{aligned}
\]
Consequently, the decrease in the optimized objective is
\[
\Delta_{j\mid S}^\lambda(\boldsymbol y)
=\boldsymbol g^\top\Omega^{-1}\boldsymbol g.
\]
Since $\widehat{\boldsymbol\beta}_{j\mid S}^\lambda(\boldsymbol y)=\Omega^{-1}\boldsymbol g=B_{j\mid S}^\lambda\boldsymbol y$, the identities in \eqref{eq:coalition-improvement} follow.
In particular, $\Delta_{j\mid S}^\lambda(\boldsymbol y)=0$ if and only if $\widehat{\boldsymbol\beta}_{j\mid S}^\lambda(\boldsymbol y)=\boldsymbol0$.
\end{proof}

\begin{lemma}[Quadratic representation of Shapley-weighted effects]
\label{lem:quadratic-shapley-effect}
For any fixed nonempty feature set $M\subseteq[m]$, target feature $j\in M$, and $\lambda>0$, the Shapley-weighted effect can be expressed as
\begin{equation}
\label{eq:shapley}
\begin{aligned}
T_{j,M}^\lambda(\boldsymbol y)
&=\boldsymbol y^\top A_{j,M}^\lambda\boldsymbol y,\\
A_{j,M}^\lambda
&=\sum_{S\subseteq M\setminus\{j\}}w_{j,M}(S)
(B_{j\mid S}^\lambda)^\top
\Omega_{j\mid S}^\lambda B_{j\mid S}^\lambda.
\end{aligned}
\end{equation}
The matrix $A_{j,M}^\lambda$ is symmetric and positive semidefinite.
\end{lemma}

\begin{proof}
Substituting \eqref{eq:coalition-improvement} into the definition of the Shapley-weighted effect \eqref{eq:shapley-loss-score} gives the stated quadratic representation.
Each summand $(B_{j\mid S}^\lambda)^\top\Omega_{j\mid S}^\lambda B_{j\mid S}^\lambda$ is symmetric and positive semidefinite because $\Omega_{j\mid S}^\lambda\succ0$.
Since every Shapley weight $w_{j,M}(S)$ is positive, $A_{j,M}^\lambda$ is symmetric and positive semidefinite.
\end{proof}

\subsection{Test-Space Characterization and Equivalent Zero-Effect Nulls}
\label{app:test-space-characterization}

We first derive an explicit representation of the test space used in Section~\ref{test-space}.
For any $\boldsymbol v\in\mathbb R^n$, the quadratic representation implies
\[
\boldsymbol v^\top A_{j,M}^\lambda\boldsymbol v
=\sum_{S\subseteq M\setminus\{j\}}w_{j,M}(S)
(B_{j\mid S}^\lambda\boldsymbol v)^\top
\Omega_{j\mid S}^\lambda
(B_{j\mid S}^\lambda\boldsymbol v).
\]
Every term is nonnegative, and each term vanishes exactly when $B_{j\mid S}^\lambda\boldsymbol v=\boldsymbol0$.
Because all weights are strictly positive, we obtain
\begin{equation}
\label{eq:appendix-kernel}
\ker(A_{j,M}^\lambda)
=\bigcap_{S\subseteq M\setminus\{j\}}\ker(B_{j\mid S}^\lambda).
\end{equation}
Taking orthogonal complements and using the symmetry of $A_{j,M}^\lambda$ gives
\[
\operatorname{range}(A_{j,M}^\lambda)
=\operatorname{span}\left\{
\operatorname{range}((B_{j\mid S}^\lambda)^\top):
S\subseteq M\setminus\{j\}\right\}.
\]
Moreover,
\[
(B_{j\mid S}^\lambda)^\top
=R_S^\lambda G_j(\Omega_{j\mid S}^\lambda)^{-1},
\]
so invertibility of $\Omega_{j\mid S}^\lambda$ yields
\[
\operatorname{range}((B_{j\mid S}^\lambda)^\top)
=\operatorname{col}(R_S^\lambda G_j).
\]
Therefore, the test space has the representation
\begin{equation}
\label{eq:target-range}
\operatorname{range}(A_{j,M}^\lambda)
=\operatorname{span}\left\{
\operatorname{col}(R_S^\lambda G_j):
S\subseteq M\setminus\{j\}\right\}.
\end{equation}

\paragraph{Equivalent coefficient-based zero-effect null.}
\label{coefficient-loss-equivalence}
An alternative measure of the contribution of $j$ is the Shapley-weighted ordinary squared norm of its fitted target coefficients:
\begin{equation}
\label{eq:weighted-coefficient-effect}
\begin{aligned}
T_{j,M}^{\mathrm{coef},\lambda}(\boldsymbol y)
&=\sum_{S\subseteq M\setminus\{j\}}w_{j,M}(S)
\|\widehat{\boldsymbol\beta}_{j\mid S}^\lambda(\boldsymbol y)\|_2^2\\
&=\boldsymbol y^\top A_{j,M}^{\mathrm{coef},\lambda}\boldsymbol y,\\
A_{j,M}^{\mathrm{coef},\lambda}
&=\sum_{S\subseteq M\setminus\{j\}}w_{j,M}(S)
(B_{j\mid S}^\lambda)^\top B_{j\mid S}^\lambda.
\end{aligned}
\end{equation}
The same positivity argument gives
\[
\ker(A_{j,M}^{\mathrm{coef},\lambda})
=\bigcap_{S\subseteq M\setminus\{j\}}\ker(B_{j\mid S}^\lambda)
=\ker(A_{j,M}^\lambda).
\]
Taking orthogonal complements and using positive semidefiniteness proves
\begin{equation}
\label{eq:equivalent-effect-null}
\begin{gathered}
T_{j,M}^\lambda(\boldsymbol\mu)=0
\quad\Longleftrightarrow\quad
T_{j,M}^{\mathrm{coef},\lambda}(\boldsymbol\mu)=0,\\
\operatorname{range}(A_{j,M}^\lambda)
=\operatorname{range}(A_{j,M}^{\mathrm{coef},\lambda}).
\end{gathered}
\end{equation}
Thus, the two effects have the same zero-effect null and the same projection-based test space, although their numerical values generally differ.

\paragraph{Expected score under a fixed target.}
For a fixed symmetric matrix $A$ and the Gaussian model \eqref{eq:model},
\[
\mathbb E_{\boldsymbol\mu}[\boldsymbol Y^\top A\boldsymbol Y]
=\boldsymbol\mu^\top A\boldsymbol\mu
+\sigma^2\operatorname{tr}(A).
\]
In particular, a positive observed loss reduction does not by itself establish a nonzero population effect.

\subsection{Proof of the Fixed-Target Distribution Theorem}
\label{app:fixed-target-distribution}

\begin{proof}[Proof of Theorem~\ref{thm:fixed-target-chi-square}]
Let $A=A_{j,M}^\lambda$, $W=\operatorname{range}(A)$, and $d=\dim(W)$, where $M$, $j$, and $\lambda$ are fixed independently of the random response $\boldsymbol Y$.
By Lemma~\ref{lem:quadratic-shapley-effect}, the matrix $A$ is symmetric and positive semidefinite.
Suppose first that $d\ge1$, and let $a_1,\ldots,a_d>0$ denote the positive eigenvalues of $A$ with corresponding orthonormal eigenvectors $\boldsymbol v_1,\ldots,\boldsymbol v_d$.
Then
\[
A=\sum_{r=1}^d a_r\boldsymbol v_r\boldsymbol v_r^\top,
\qquad
P_W=\sum_{r=1}^d\boldsymbol v_r\boldsymbol v_r^\top.
\]
Under the zero-effect null, $T_{j,M}^\lambda(\boldsymbol\mu)=0$, and the quadratic representation gives
\[
0=\boldsymbol\mu^\top A\boldsymbol\mu
=\sum_{r=1}^d a_r(\boldsymbol v_r^\top\boldsymbol\mu)^2.
\]
Because all $a_r$ are strictly positive, $\boldsymbol v_r^\top\boldsymbol\mu=0$ for every $r\in[d]$, which also proves $P_W\boldsymbol\mu=\boldsymbol0$.
Define
\[
Z_r=\frac{\boldsymbol v_r^\top\boldsymbol Y}{\sigma},
\qquad r\in[d].
\]
Since the eigenvectors are fixed and orthonormal and $\boldsymbol Y\sim\mathcal N(\boldsymbol\mu,\sigma^2 I_n)$, the variables $Z_1,\ldots,Z_d$ are independent standard normal random variables under the null.
The spectral decomposition therefore gives
\[
\frac{T_{j,M}^\lambda(\boldsymbol Y)}{\sigma^2}
=\frac{\boldsymbol Y^\top A\boldsymbol Y}{\sigma^2}
=\sum_{r=1}^d a_rZ_r^2,
\]
and
\[
Q_{j,M}^\lambda(\boldsymbol Y)
=\frac{\|P_W\boldsymbol Y\|_2^2}{\sigma^2}
=\sum_{r=1}^d Z_r^2
\sim\chi_d^2.
\]
If $d=0$, then $A=0$ and $P_W=0$, so both statistics are identically zero.
This completes the proof.
\end{proof}

\subsection{Numerical Test Space}
\label{app:numerical-test-space-section3}
\label{numerical-target}

In numerical implementations, a prespecified tolerance $\tau>0$ may be used for basis compression and eigenvalue truncation, giving a retained test space $W_{j,\tau}$.
The corresponding null is
\begin{equation}
\label{eq:numerical-null}
H_{0,j,M_{\rm obs}}^{(\tau)}:
P_{W_{j,\tau}}\boldsymbol\mu=\boldsymbol0.
\end{equation}
This null is identical to the ideal null only when $W_{j,\tau}=W_j$; numerical truncation can change the inferential target.
The construction and its use in the selection procedure are described in Appendix~\ref{appendix:numerical-implementation}.

\section{Proof of Theorem~\ref{thm:selective-validity}}
\label{appendix:proof-selective-validity}

\begin{proof}[Proof of Theorem~\ref{thm:selective-validity}]
Fix the design matrix $X$, the prespecified basis functions and cross-validation folds, and possible observed selection labels $M_{\rm obs}$ and $\lambda_{\rm obs}$.
We also fix a target feature $j\in M_{\rm obs}$ whose choice involves no additional response-dependent selection beyond the event $\mathcal E_{\rm obs}$.
For these fixed labels, define the effect matrix $A_{j,M_{\rm obs}}^{\lambda_{\rm obs}}$, the test space $W_j$, and its orthogonal projector $P=P_{W_j}$; these are deterministic for the Gaussian calculation before imposing the selection event.
Set $d=d_j=\operatorname{rank}(P)\ge1$ and consider the fixed-label null hypothesis $P\boldsymbol\mu=\boldsymbol0$.

\paragraph{Distribution of the radius before selection.}
Let $V\in\mathbb R^{n\times d}$ have orthonormal columns spanning $W_j$, so that $V^\top V=I_d$ and $P=VV^\top$.
Decompose the response as
\begin{equation}
\label{eq:appendix-si-decomposition}
\boldsymbol Y=V\boldsymbol\xi+\boldsymbol Z_j,
\qquad
\boldsymbol\xi=V^\top\boldsymbol Y,
\qquad
\boldsymbol Z_j=(I_n-P)\boldsymbol Y.
\end{equation}
Under the Gaussian model \eqref{eq:model} and the null $P\boldsymbol\mu=\boldsymbol0$, we have
\[
\boldsymbol\xi\sim\mathcal N(\boldsymbol0,\sigma^2 I_d),
\qquad
\operatorname{Cov}(\boldsymbol\xi,\boldsymbol Z_j)
=\sigma^2V^\top(I_n-P)=0.
\]
Since $\boldsymbol\xi$ and $\boldsymbol Z_j$ are jointly Gaussian, the zero cross-covariance implies their independence.
Define
\[
R_j(\boldsymbol Y)=\frac{\|\boldsymbol\xi\|_2}{\sigma},
\qquad
\boldsymbol D_j=\frac{V\boldsymbol\xi}{\|\boldsymbol\xi\|_2}.
\]
These definitions agree with \eqref{eq:radial-statistic} because $P\boldsymbol Y=V\boldsymbol\xi$ and $V$ has orthonormal columns.
By the spherical symmetry of $\boldsymbol\xi/\sigma\sim\mathcal N(\boldsymbol0,I_d)$, its norm has the $\chi_d$ distribution and is independent of its direction.
Moreover, the independence of $\boldsymbol\xi$ and $\boldsymbol Z_j$ implies that $R_j(\boldsymbol Y)$ is jointly independent of $(\boldsymbol Z_j,\boldsymbol D_j)$.
Consequently, for almost every value $(\boldsymbol z_j,\boldsymbol u_j)$ of the orthogonal component and direction, the conditional density of the radius before imposing selection is $f_{\chi_d}(r)$ for $r\ge0$.

\paragraph{Conditioning on the selection event.}
Because $d\ge1$, the event $R_j(\boldsymbol Y)=0$ has probability zero, so the direction is almost surely well defined.
For fixed values $\boldsymbol Z_j=\boldsymbol z_j$ and $\boldsymbol D_j=\boldsymbol u_j$, the response admits the one-dimensional representation
\begin{equation}
\label{eq:appendix-si-ray}
\boldsymbol Y=\boldsymbol z_j+\sigma R_j(\boldsymbol Y)\boldsymbol u_j
=\boldsymbol y_j\bigl(R_j(\boldsymbol Y)\bigr).
\end{equation}
By \eqref{eq:selection-event} and \eqref{eq:selection-region}, on this ray the event $\mathcal E_{\rm obs}$ is equivalent to the condition $R_j(\boldsymbol Y)\in\mathcal Z$, where
\[
\mathcal Z
=\left\{r\ge0:
M(\boldsymbol y_j(r))=M_{\rm obs},\quad
\widehat\lambda(\boldsymbol y_j(r))=\lambda_{\rm obs}
\right\}.
\]
The set $\mathcal Z$ is determined by the fixed conditioning values $(M_{\rm obs},\lambda_{\rm obs},\boldsymbol z_j,\boldsymbol u_j)$.
For any Borel set $B\subseteq[0,\infty)$ and any nondegenerate conditioning values for which $\int_{\mathcal Z}f_{\chi_d}(t)\,dt>0$, independence of the radius from $(\boldsymbol Z_j,\boldsymbol D_j)$ gives
\begin{equation}
\label{eq:appendix-si-conditional-probability}
\begin{aligned}
&\mathbb P_{\boldsymbol\mu}\!\left(
R_j(\boldsymbol Y)\in B\mid
\mathcal E_{\rm obs},\,
\boldsymbol Z_j=\boldsymbol z_j,\,
\boldsymbol D_j=\boldsymbol u_j
\right)\\
&\qquad=
\frac{\displaystyle\int_{B\cap\mathcal Z}f_{\chi_d}(t)\,dt}
{\displaystyle\int_{\mathcal Z}f_{\chi_d}(t)\,dt}.
\end{aligned}
\end{equation}
This establishes the truncated-chi density in \eqref{eq:conditional-density}.

\paragraph{Conditional validity of the selective $p$-value.}
Let $R$ denote a random variable with the conditional distribution in \eqref{eq:appendix-si-conditional-probability}, and define its conditional survival function by
\begin{equation}
\label{eq:appendix-si-survival}
S_{\mathcal Z}(r)
=
\frac{\displaystyle\int_{\mathcal Z\cap[r,\infty)}f_{\chi_d}(t)\,dt}
{\displaystyle\int_{\mathcal Z}f_{\chi_d}(t)\,dt}.
\end{equation}
The conditional distribution of $R$ is absolutely continuous, even when $\mathcal Z$ consists of several disjoint intervals.
Therefore, the probability integral transform gives $S_{\mathcal Z}(R)\sim\operatorname{Uniform}(0,1)$ under the fixed-label null.
On the conditioned ray, the selective $p$-value in \eqref{eq:selective-pvalue} is precisely $p_j^{\rm SI}(\boldsymbol Y)=S_{\mathcal Z}(R_j(\boldsymbol Y))$, with observed value $p_j^{\rm SI}=S_{\mathcal Z}(r_{\rm obs})$.
It follows that, for every $\alpha\in(0,1)$ and almost every admissible conditioning value,
\begin{equation}
\label{eq:appendix-si-uniform-validity}
\begin{aligned}
&\mathbb P_{\boldsymbol\mu}\!\left(
 p_j^{\rm SI}(\boldsymbol Y)\le\alpha\mid
 \mathcal E_{\rm obs},\,
 \boldsymbol Z_j=\boldsymbol z_j,\,
 \boldsymbol D_j=\boldsymbol u_j
 \right)\\
&\qquad=\alpha\le\alpha.
\end{aligned}
\end{equation}
This proves the conditional type I error guarantee in \eqref{eq:conditional-validity}.
The assertion is conditional on a fixed selected target and its associated null hypothesis; any further response-dependent rule for choosing $j$ must also be included in the selection event.
\end{proof}

\section{Test Space and Global Null}\label{appendix:test-space-global-null}

Fix a nonempty feature set \(M\subseteq[m]\), a target feature \(j\in M\), and a regularization parameter \(\lambda>0\).
Write \(W_j=\operatorname{range}(A_{j,M}^\lambda)\).
By \eqref{eq:target-range} in Appendix~\ref{app:test-space-characterization},
\begin{equation}\label{eq:ridge-test-space-range}
W_j
=\operatorname{range}(A_{j,M}^\lambda)
=\operatorname{span}\left\{
\operatorname{col}(R_S^\lambda G_j):
S\subseteq M\setminus\{j\}
\right\},
\end{equation}
where \(R_S^\lambda=C_0-H_S^\lambda\) and \(H_S^\lambda=K_S(K_S+n\lambda I_n)^{-1}\) for \(K_S=G_SG_S^\top\) (with \(H_\varnothing^\lambda=0\)).

Define the combined centered basis subspace associated with \(M\) by
\[
\mathcal V_M
=\operatorname{span}\left\{
\operatorname{col}(G_S):
\varnothing\neq S\subseteq M
\right\}.
\]
Because every column of \(G_j\) is centered, \(C_0G_j=G_j\), and hence
\[
R_S^\lambda G_j = G_j - H_S^\lambda G_j
\]
for every \(S\subseteq M\setminus\{j\}\).
Every column of \(G_j\) lies in \(\operatorname{col}(G_{\{j\}})\subseteq\mathcal V_M\), and the range of \(H_S^\lambda\) is contained in \(\operatorname{col}(G_S)\subseteq\mathcal V_M\).
Thus \(W_j\subseteq\mathcal V_M\).
In fact, when \(|M|\ge2\), only proper subsets \(S\subsetneq M\) appear in \eqref{eq:ridge-test-space-range}, so \(W_j\) is contained in the proper-subset span \(\mathcal V_M^{\mathrm{prop}}=\operatorname{span}\{\operatorname{col}(G_S):\varnothing\neq S\subsetneq M\}\subseteq\mathcal V_M\).

Together with \eqref{eq:projection-null}, the inclusion \(W_j\subseteq\mathcal V_M\) implies that the global null \(P_{\mathcal V_M}\boldsymbol\mu=\boldsymbol0\) within \(\mathcal V_M\) implies the feature's zero-effect null \(P_{W_j}\boldsymbol\mu=\boldsymbol0\).
The two nulls are equivalent if and only if \(W_j=\mathcal V_M\).
Indeed, if the inclusion is strict, any nonzero \(\boldsymbol\mu\in\mathcal V_M\cap W_j^\perp\) satisfies \(P_{W_j}\boldsymbol\mu=\boldsymbol0\) while \(P_{\mathcal V_M}\boldsymbol\mu=\boldsymbol\mu\ne\boldsymbol0\).

As a simple illustration, consider \(M=\{1,2\}\) with single-column centered univariate basis blocks \(G_1=[\boldsymbol g_1]\) and \(G_2=[\boldsymbol g_2]\) such that \(\boldsymbol g_1,\boldsymbol g_2\ne\boldsymbol0\) and \(\boldsymbol g_1^\top\boldsymbol g_2\ne0\).
For feature \(1\), the nuisance coalitions are \(S=\varnothing\) and \(S=\{2\}\), with \(H_\varnothing^\lambda=0\) and
\[
H_{\{2\}}^\lambda
=\frac{\boldsymbol g_2\boldsymbol g_2^\top}
{\|\boldsymbol g_2\|_2^2+n\lambda}.
\]
Applying \eqref{eq:ridge-test-space-range} gives
\[
W_1
=\operatorname{span}\left\{
\boldsymbol g_1,\;
\boldsymbol g_1-\frac{\boldsymbol g_2^\top\boldsymbol g_1}
{\|\boldsymbol g_2\|_2^2+n\lambda}\boldsymbol g_2
\right\}
=\operatorname{span}\{\boldsymbol g_1,\boldsymbol g_2\}
=\mathcal V_{\{1,2\}}^{\mathrm{prop}},
\]
because nonorthogonality makes the coefficient of \(\boldsymbol g_2\) nonzero.
Interchanging the feature indices yields the same test space \(W_2=W_1\).
Thus, when two nonorthogonal features are tested within \(M_{\rm obs}=\{1,2\}\), they can share the same zero-effect null even though their Shapley-weighted loss-reduction scores differ.

The same subspace relation applies to the compressed basis blocks \(\widetilde G_S=QQ^\top G_S\) in Appendix~\ref{appendix:numerical-implementation} before target eigenvectors are truncated.
Subsequent eigenvalue truncation can strictly reduce the test space, so the implemented null \eqref{eq:numerical-null} coincides with the global null within the compressed basis span only when \(W_{j,\tau}\) equals that span.

\section{Numerical Implementation}\label{appendix:numerical-implementation}

To handle near-collinear coalition basis blocks stably, we compute the singular value decomposition of the horizontally concatenated nonempty centered blocks \(G_S\) (\(\varnothing\neq S\subseteq[m]\)) from \eqref{eq:basis} and collect the left singular vectors whose singular values exceed \(\tau\max(1,s_{\max})\) as the orthonormal columns of \(Q\in\mathbb R^{n\times q}\), where \(\tau>0\) is a relative rank tolerance (\(\tau=10^{-10}\) in our experiments) and \(s_{\max}\) is the largest singular value.
We then form the selection and target matrices in the \(q\)-dimensional \(Q\)-coordinates using \(Q^\top G_S\in\mathbb R^{q\times p_S}\), while leaving the cross-validation matrices \(C_{\lambda_\ell}\) in \eqref{eq:cv} unchanged on \(\mathbb R^n\).
On \(\mathbb R^n\), this is equivalent to replacing each \(G_S\) by its projection \(\widetilde G_S = QQ^\top G_S\), which leaves the selection and target matrices exact whenever \(Q\) spans the full column space of \(\{G_S\}\).

Because near-zero eigenvalues distort the target rank \(d_j\) (the \(\chi_{d_j}\) degrees of freedom), we retain the orthonormal eigenvectors of the \(q\times q\) reduced target matrix \(Q^\top \widetilde A_{j,M_{\rm obs}}^{\lambda_{\rm obs}} Q\) for \(j\in M_{\rm obs}\) whose eigenvalues exceed \(\tau a_{\max}\), where \(\widetilde A_{j,M_{\rm obs}}^{\lambda_{\rm obs}}\) is the compressed target matrix built from \(\{\widetilde G_S\}\) and \(a_{\max}\) is its largest eigenvalue, as columns of \(B\in\mathbb R^{q\times d_{j,\tau}}\) and set \(W_{j,\tau}=\operatorname{col}(QB)\), returning \(p_j^{\rm SI}=1\) if \(Q\) or \(B\) is empty.
Discarding a small positive eigenvalue can shrink the test space, so \(\tau\) is part of the implemented null \eqref{eq:numerical-null}; selective validity in Section~\ref{selective-inference} still holds when the same compressed scores define the selection event.

To prevent spurious roots and upper-tail cancellation in \eqref{eq:selective-pvalue}, coefficients in \eqref{eq:quadratic-comparison} below a machine-precision threshold (distinct from \(\tau\)) are zeroed before root-finding, and chi interval probabilities are evaluated on the log scale using upper-tail survival functions and log-sum-exp, clamping \(p_j^{\rm SI}\) to \([0,1]\) and returning \(1\) if \(r_{\rm obs}=0\).
These safeguards preserve \(p_j^{\rm SI}\) in exact arithmetic.

\section{Details of the Experiments}\label{appendix:experiments}

\subsection{Experimental Setup}\label{appendix:experimental-setup}

This subsection provides the experimental details omitted from Section~\ref{numerical-experiments}.

\paragraph{Synthetic data and trial protocol.}
Each synthetic trial generates independent rows of \(X\) from \(\mathcal N(\boldsymbol0,\Gamma)\) with \(\Gamma\in\mathbb R^{m\times m}\) and \(\Gamma_{ab}=\rho^{|a-b|}\) (using \(0^0:=1\)), and draws the design, basis centers, cross-validation folds, response noise, and tested feature \(J\in M(\boldsymbol Y)\) anew.
We run 10,000 trials per configuration; corresponding trials across signal functions and signal strengths share the design, noise draw, basis and fold choices, and target randomization, and samples are nested across \(n\in\{20,40,60\}\).
The \(n=40\) baseline null trials are reused in the basis-sensitivity sweep.
Conditional rejection rates are estimated from trials with \(J=1\); overall null rejection rates across all 10,000 trials are summarized in Appendix~\ref{appendix:additional-null}.

\paragraph{Basis construction.}
For each nonempty feature subset \(S\subseteq[m]\), we construct \(p_S=\lceil\eta n\rceil\) Gaussian radial basis functions (with default center fraction \(\eta=0.1\)), \(\phi_{S,q}(\boldsymbol x)=\exp\bigl(-\tfrac12\|\boldsymbol x-\boldsymbol c_{S,q}\|_2^2/h_S^2\bigr)\), and column-center them as in \eqref{eq:basis}.
The centers \(\boldsymbol c_{S,q}\) are the feature vectors \(X_{i,S}\) at \(p_S\) row indices sampled without replacement from \(\{1,\ldots,n\}\) (shared across subsets on the full data).
The bandwidth \(h_S\) is the median pairwise Euclidean distance among the rows of \(X_{\cdot,S}\); if this median is zero (for discrete or constant features), we use the median of the positive pairwise distances, or \(1\) if all pairwise distances are zero.
With \(\eta=0.1\), the full-data numbers of centers per basis are 2, 4, and 6 at \(n=20,40,60\), respectively, and the fold-local numbers are 2, 4, and 5; increasing \(n\) therefore also changes the basis size and the specified test space.
Within each CV fold, centers, bandwidths, and centering means are computed using only the training rows.
All methods use the numerical test space \(W_{j,\tau}\) with \(\tau=10^{-10}\) in \eqref{eq:numerical-null}.

\paragraph{Signal functions.}
For the first feature, the four signal functions used in Section~\ref{experiments-power} and Appendix~\ref{appendix:additional-power} are
\[
f_{\mathrm{lin}}(t)=t,\qquad
f_{\mathrm{quad}}(t)=t^2-1,\qquad
f_{\mathrm{mix}}(t)=t+\tfrac12(t^2-1),\qquad
f_{\mathrm{sin}}(t)=\sin(t).
\]
For each realized \(X\) and signal function \(f\), let \(\boldsymbol f=(f(X_{11}),\ldots,f(X_{n1}))^\top\) and \(\bar f=n^{-1}\sum_{i=1}^n f(X_{i1})\).
We standardize the signal vector to unit root mean square (RMS) via
\[
\widetilde{\boldsymbol f}
=\frac{\boldsymbol f-\bar f\mathbf1_n}
{\{n^{-1}\|\boldsymbol f-\bar f\mathbf1_n\|_2^2\}^{1/2}},
\]
so that \(\delta\) equals the realized signal RMS divided by the noise standard deviation \(\sigma=1\).
All synthetic signals are univariate additive functions of the first feature.

\subsection{Methods for Comparison}\label{appendix:comparison-methods}

This subsection formalizes the baseline methods compared in Section~\ref{numerical-experiments}.
Let \(r_{\rm obs}\) and \(d_j\) denote the observed radius and retained rank for the tested feature \(j\in M_{\rm obs}\).
For zero-rank targets (\(d_j=0\)), we set the \(p\)-value to \(1\) for every method.
Otherwise, the naive \(p\)-value is
\[
p_j^{\rm naive}=\mathbb P(\chi_{d_j}\ge r_{\rm obs}),
\]
Proposed uses the selective \(p\)-value \(p_j^{\rm SI}\) in \eqref{eq:selective-pvalue}, incorporating both the feature-selection comparisons \eqref{eq:topk-comparison} and the cross-validation comparisons \eqref{eq:cv-comparison}, and Bonferroni uses
\[
p_j^{\rm Bonf}
=\min\{1,N_{\rm Bonf}\,p_j^{\rm naive}\},\qquad
N_{\rm Bonf}=Lk\binom{m}{k},
\]
accounting for all \(L\binom{m}{k}\) regularization-parameter and selected-set combinations and \(k\) tested features per combination.
Note that Bonferroni controls familywise error over the fixed-label candidate tests rather than error conditional on the observed selected set and regularization parameter.

\subsection{Additional Null Results}\label{appendix:additional-null}

This subsection complements Figure~\ref{fig:null-main} with overall null rejection rates, basis-size sensitivity, and results stratified by the cross-validated penalty.
Across all ten null settings (10,000 trials each; 974--1,035 trials with \(J=1\)), overall rejection rates at \(\alpha=0.05\) range from 0.0462 to 0.0519 for Proposed (conditional rates 0.0451--0.0578), from 0.1400 to 0.2610 for Naive, and are at most 0.0004 for Bonferroni.

\begin{figure}[!htbp]
\centering
\includegraphics[width=\linewidth]{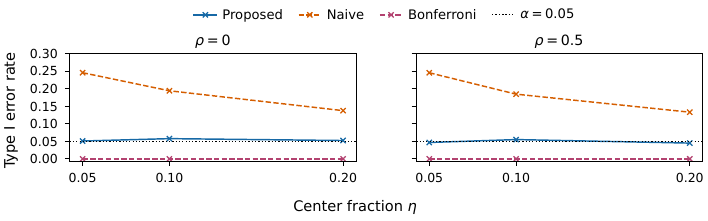}
\caption{Type I error rate conditional on \(J=1\) at \(n=40\) as the RBF center fraction \(\eta\) varies.
Proposed remains near the nominal level across all values of \(\eta\).}\label{fig:sec6-basis-sensitivity}\label{fig:basis-sensitivity}
\end{figure}

\paragraph{Basis sensitivity.}
To assess sensitivity to the number of RBF centers, we vary \(\eta\in\{0.05,0.1,0.2\}\) at \(n=40\) while keeping all other settings fixed.
Figure~\ref{fig:sec6-basis-sensitivity} shows that Proposed remains near the nominal level \(\alpha=0.05\) across all values of \(\eta\) (conditional rates 0.045--0.058), whereas Naive is anti-conservative (0.133--0.246) and Bonferroni makes no rejections among tests with \(J=1\).

\begin{figure}[!htbp]\centering
\includegraphics[width=\linewidth]{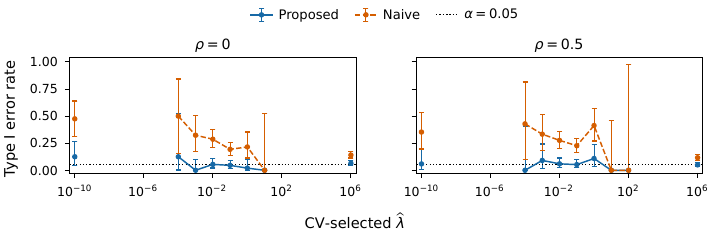}
\caption{Type I error rate conditional on \(J=1\) within CV-selected penalty groups at \(n=40\).
Error bars show pointwise 95\% exact binomial intervals.}
\label{fig:lambda-main}\end{figure}

\paragraph{Selected-penalty groups.}
Figure~\ref{fig:lambda-main} reports conditional null rejection rates stratified by the CV-selected penalty \(\widehat\lambda\) at \(n=40\) and \(\eta=0.1\), with pointwise 95\% Clopper--Pearson intervals.
Across well-populated penalty groups, Proposed remains close to the nominal level \(\alpha=0.05\), whereas Naive is consistently anti-conservative; sparsely selected penalties have wide intervals due to small group counts.

\subsection{Additional Power Results}\label{appendix:additional-power}

\begin{figure}[!htbp]\centering
\includegraphics[width=\linewidth]{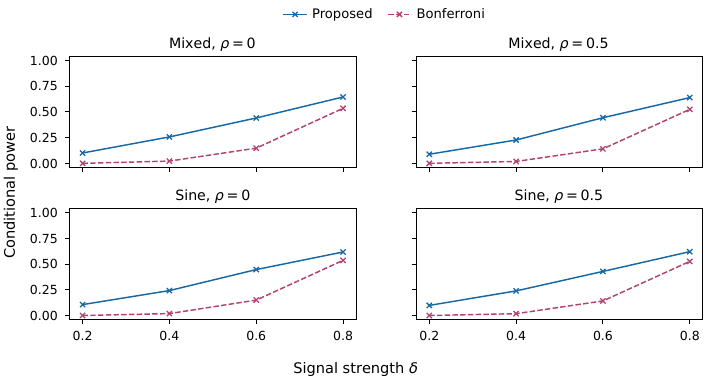}
\caption{Conditional power given \(J=1\) at \(n=40\) for the mixed and sine signals (see Figure~\ref{fig:power-selected} for linear and quadratic signals).}
\label{fig:power-selected-40}\end{figure}

\begin{figure}[!htbp]\centering
\includegraphics[width=\linewidth]{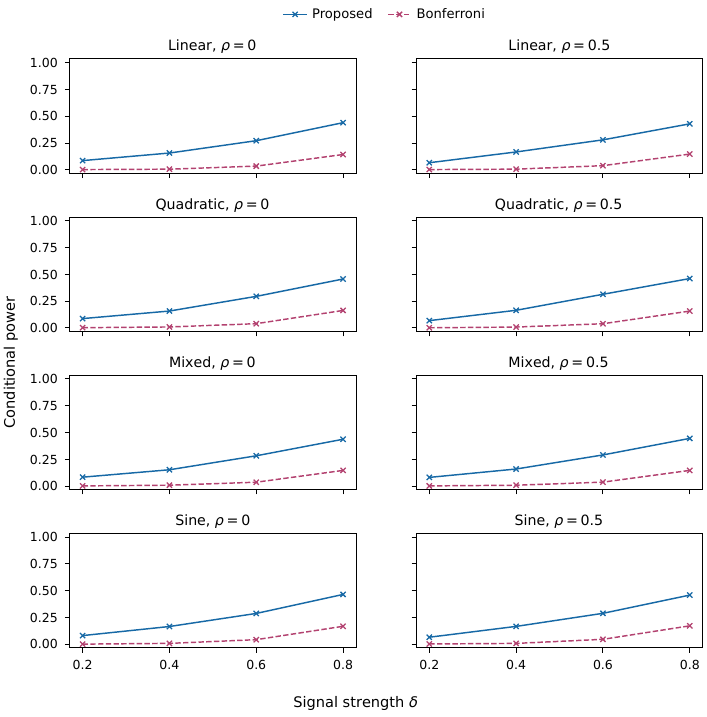}
\caption{Conditional power given \(J=1\) at \(n=20\) as the signal strength \(\delta\) varies.}
\label{fig:power-selected-20}\end{figure}

\begin{figure}[!htbp]\centering
\includegraphics[width=\linewidth]{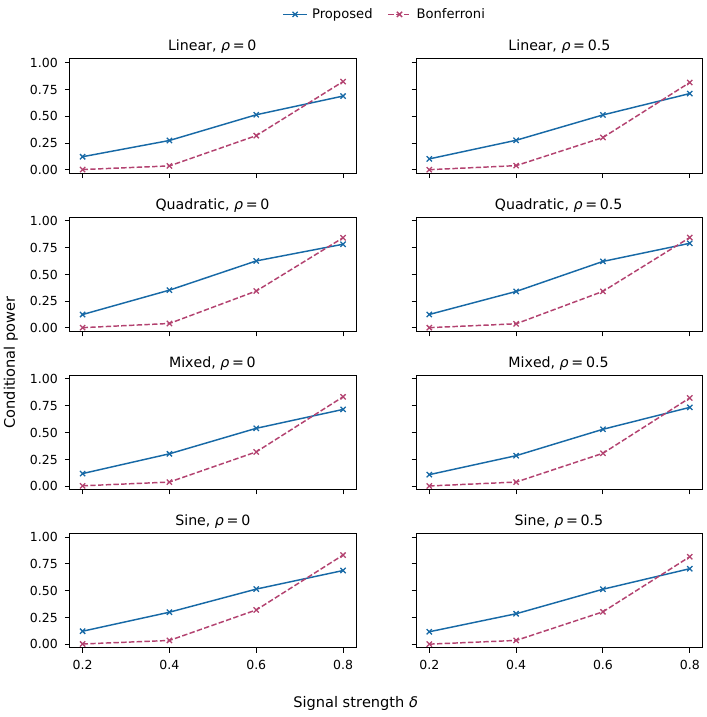}
\caption{Conditional power given \(J=1\) at \(n=60\) as the signal strength \(\delta\) varies.}
\label{fig:power-selected-60}\end{figure}

Together with Figure~\ref{fig:power-selected}, Figures~\ref{fig:power-selected-40}--\ref{fig:power-selected-60} report conditional power given \(J=1\) for Proposed and Bonferroni across all four signal functions at \(n\in\{20,40,60\}\).
Each configuration uses 10,000 datasets, with shared conditional denominators (\(J=1\)) ranging from 1,343 to 5,011 across positive-signal settings.
Proposed achieves higher conditional power than Bonferroni across all settings at \(n=20\) and \(n=40\), and for weak-to-moderate signals at \(n=60\); at the strongest signal and largest sample size (\(\delta=0.8\), \(n=60\)), Bonferroni slightly exceeds Proposed across the eight signal-shape and correlation configurations.

\clearpage
\subsection{Real Data Details}\label{appendix:real-data-details}

\paragraph{Datasets and preprocessing.}
We evaluate eight regression benchmarks from the UCI Machine Learning Repository (all licensed under CC BY 4.0), summarized in Table~\ref{tab:real-datasets}.
Prior to sub-sampling, all predictors and responses are standardized to zero mean and unit variance.
For any predictor with fewer than \(0.3N\) distinct values (where \(N\) is the original dataset size), independent Gaussian jitter with standard deviation \(0.1\) is added once during preprocessing to avoid degenerate pairwise distances.

\begin{table}[!htbp]\centering\footnotesize
\caption{Real datasets used in Section~\ref{experiments-real-data}, with original sample size \(N\) and number of retained predictors \(m\) (TEY denotes turbine energy yield).}\label{tab:real-datasets}
\vspace{5pt}\setlength{\tabcolsep}{4pt}
\begin{tabular*}{\linewidth}{@{\extracolsep{\fill}}llrrll@{}}\hline
Label & Dataset & \(N\) & \(m\) & Response & Notes\\\hline
Data1 & Abalone~\cite{nash1994abalone} & 4,177 & 7 & Rings & Excl.\ sex\\
Data2 & Concrete~\cite{yeh1998concrete} & 1,030 & 8 & Strength & --\\
Data3 & Energy Efficiency~\cite{tsanas2012energy} & 768 & 8 & Cooling load & --\\
Data4 & Gas Turbine~\cite{gas2019emission} & 7,384 & 10 & TEY & 2015 subset\\
Data5 & Energy Efficiency~\cite{tsanas2012energy} & 768 & 8 & Heating load & --\\
Data6 & Real Estate~\cite{yeh2018realestate} & 414 & 7 & Unit price & Incl.\ index col.\\
Data7 & Wine Quality (Red)~\cite{cortez2009wine} & 1,599 & 11 & Quality & --\\
Data8 & Wine Quality (White)~\cite{cortez2009wine} & 4,898 & 11 & Quality & --\\
\hline\end{tabular*}\end{table}

\paragraph{Experimental protocol.}
For each dataset and sample size \(n\in\{20,40,60\}\), we draw 10,000 sub-samples of size \(n\) without replacement (nested across \(n\) within corresponding trials), adding no synthetic signal or noise to the observed responses.
Both Proposed and Bonferroni are evaluated on the exact same sub-samples, RBF bases (constructed per sub-sample, and fold-locally for CV), folds, selected pairs, tested features, and test spaces.
Finally, \(\sigma=1\) is treated as a working assumption for the standardized response rather than a known or independently estimated residual standard deviation; accordingly, the empirical power interpretation in Section~\ref{experiments-real-data} relies on the relevance and noise assumptions and does not verify real-data type I error control.

\subsection{Computational Resources}\label{appendix:computational-resources}

We executed all numerical experiments on dual Intel Xeon Gold 6338 processors with 64 cores (2.00 GHz) and 512 GB of memory.

\IfFileExists{unsrturl.bst}{\bibliographystyle{unsrturl}}{\bibliographystyle{unsrt}}
\bibliography{ref}

@article{fithian2014optimal,
  title={Optimal inference after model selection},
  author={Fithian, William and Sun, Dennis and Taylor, Jonathan},
  journal={arXiv preprint arXiv:1410.2597},
  year={2014}
}

@article{taylor2015statistical,
  title={Statistical learning and selective inference},
  author={Taylor, Jonathan and Tibshirani, Robert J},
  journal={Proceedings of the National Academy of Sciences},
  volume={112},
  number={25},
  pages={7629--7634},
  year={2015},
  publisher={National Academy of Sciences},
  doi={10.1073/pnas.1507583112}
}

@article{lee2016exact,
  title={Exact post-selection inference, with application to the lasso},
  author={Lee, Jason D and Sun, Dennis L and Sun, Yuekai and Taylor, Jonathan E},
  journal={The Annals of Statistics},
  volume={44},
  number={3},
  pages={907--927},
  year={2016},
  publisher={Institute of Mathematical Statistics},
  doi={10.1214/15-AOS1371}
}

@article{tibshirani2016exact,
  title={Exact post-selection inference for sequential regression procedures},
  author={Tibshirani, Ryan J and Taylor, Jonathan and Lockhart, Richard and Tibshirani, Robert},
  journal={Journal of the American Statistical Association},
  volume={111},
  number={514},
  pages={600--620},
  year={2016},
  publisher={Taylor \& Francis},
  doi={10.1080/01621459.2015.1108848}
}

@article{rugamer2020inference,
title={Inference for {L}2-boosting},
author={R{\"u}gamer, David and Greven, Sonja},
journal={Statistics and Computing},
volume={30},
number={2},
pages={279--289},
year={2020},
publisher={Springer}
}

@inproceedings{yamada2018post,
title={Post selection inference with kernels},
author={Yamada, Makoto and Umezu, Yuta and Fukumizu, Kenji and Takeuchi, Ichiro},
booktitle={International Conference on Artificial Intelligence and Statistics},
pages={152--160},
year={2018},
organization={PMLR}
}

@article{gao2024selective,
  title={Selective inference for hierarchical clustering},
  author={Gao, Lucy L. and Bien, Jacob and Witten, Daniela},
  journal={Journal of the American Statistical Association},
  volume={119},
  number={545},
  pages={332--342},
  year={2024},
  publisher={Taylor \& Francis},
  doi={10.1080/01621459.2022.2116331}
}

@inproceedings{duy2020computing,
  title={Computing Valid p-value for Optimal Changepoint by Selective Inference using Dynamic Programming},
  author={Duy, Vo Nguyen Le and Toda, Hiroki and Sugiyama, Ryota and Takeuchi, Ichiro},
  booktitle={Advances in Neural Information Processing Systems},
  volume={33},
  pages={11356--11367},
  year={2020},
  publisher={Curran Associates, Inc.}
}

@article{jewell2022testing,
  title={Testing for a change in mean after changepoint detection},
  author={Jewell, Sean and Fearnhead, Paul and Witten, Daniela},
  journal={Journal of the Royal Statistical Society Series B: Statistical Methodology},
  volume={84},
  number={4},
  pages={1082--1104},
  year={2022},
  doi={10.1111/rssb.12501}
}

@article{yamada2025change,
  title={Change Point Detection in the Frequency Domain with Statistical Reliability},
  author={Yamada, Akifumi and Shiraishi, Tomohiro and Nishino, Shuichi and Katsuoka, Teruyuki and Taji, Kouichi and Takeuchi, Ichiro},
  journal={Transactions on Machine Learning Research},
  issn={2835-8856},
  year={2025},
  url={https://openreview.net/forum?id=FNRdaHz3qN}
}

@article{tsukurimichi2022conditional,
  title={Conditional selective inference for robust regression and outlier detection using piecewise-linear homotopy continuation},
  author={Tsukurimichi, Toshiaki and Inatsu, Yu and Duy, Vo Nguyen Le and Takeuchi, Ichiro},
  journal={Annals of the Institute of Statistical Mathematics},
  volume={74},
  pages={1197--1228},
  year={2022},
  publisher={Springer},
  doi={10.1007/s10463-022-00846-2}
}

@inproceedings{le2021parametric,
  title={Parametric Programming Approach for More Powerful and General Lasso Selective Inference},
  author={Nguyen Le Duy, Vo and Takeuchi, Ichiro},
  booktitle={Proceedings of The 24th International Conference on Artificial Intelligence and Statistics},
  pages={901--909},
  year={2021},
  editor={Banerjee, Arindam and Fukumizu, Kenji},
  volume={130},
  series={Proceedings of Machine Learning Research},
  publisher={PMLR},
  url={https://proceedings.mlr.press/v130/nguyen-le-duy21a.html}
}

@inproceedings{sugiyama2021more,
  title={More Powerful and General Selective Inference for Stepwise Feature Selection using Homotopy Method},
  author={Sugiyama, Kazuya and Duy, Vo Nguyen Le and Takeuchi, Ichiro},
  booktitle={Proceedings of the 38th International Conference on Machine Learning},
  pages={9891--9901},
  year={2021},
  editor={Meila, Marina and Zhang, Tong},
  volume={139},
  series={Proceedings of Machine Learning Research},
  publisher={PMLR},
  url={https://proceedings.mlr.press/v139/sugiyama21a.html}
}

@inproceedings{duy2022quantifying,
  title={Quantifying Statistical Significance of Neural Network-based Image Segmentation by Selective Inference},
  author={Duy, Vo Nguyen Le and Iwazaki, Shogo and Takeuchi, Ichiro},
  booktitle={Advances in Neural Information Processing Systems},
  volume={35},
  pages={31627--31639},
  year={2022},
  publisher={Curran Associates, Inc.},
  doi={10.52202/068431-2293}
}

@inproceedings{miwa2023salient,
  title={Valid P-Value for Deep Learning-Driven Salient Region},
  author={Miwa, Daiki and Duy, Vo Nguyen Le and Takeuchi, Ichiro},
  booktitle={The Eleventh International Conference on Learning Representations},
  year={2023},
  url={https://openreview.net/forum?id=qihMOPw4Sf_}
}

@inproceedings{niihori2025quantifying,
  title={Quantifying Statistical Significance of Deep Nearest Neighbor Anomaly Detection via Selective Inference},
  author={Niihori, Mizuki and Nishino, Shuichi and Katsuoka, Teruyuki and Shiraishi, Tomohiro and Taji, Kouichi and Takeuchi, Ichiro},
  booktitle={Advances in Neural Information Processing Systems},
  volume={38},
  pages={173174--173203},
  year={2025},
  publisher={Curran Associates, Inc.},
  doi={10.52202/085713-5762}
}

@inproceedings{shiraishi2024statistical,
  title={Statistical Test for Attention Maps in Vision Transformers},
  author={Shiraishi, Tomohiro and Miwa, Daiki and Katsuoka, Teruyuki and Duy, Vo Nguyen Le and Taji, Kouichi and Takeuchi, Ichiro},
  booktitle={Proceedings of the 41st International Conference on Machine Learning},
  pages={45079--45096},
  year={2024},
  volume={235},
  series={Proceedings of Machine Learning Research},
  publisher={PMLR},
  url={https://proceedings.mlr.press/v235/shiraishi24a.html}
}

@article{shiraishi2026statistical,
  title={Statistical Test for Attention in Transformers for Images and Time Series},
  author={Shiraishi, Tomohiro and Miwa, Daiki and Katsuoka, Teruyuki and Duy, Vo Nguyen Le and Nishino, Shuichi and Taji, Kouichi and Takeuchi, Ichiro},
  journal={Journal of Machine Learning Research},
  volume={27},
  number={119},
  pages={1--43},
  year={2026}
}

@incollection{shapley1953value,
  author = {Shapley, Lloyd S.},
  title = {A Value for {n}-Person Games},
  booktitle = {Contributions to the Theory of Games, Volume II},
  editor = {Kuhn, Harold W. and Tucker, Albert W.},
  pages = {307--317},
  publisher = {Princeton University Press},
  year = {1953}
}

@article{strumbelj2010efficient,
  author = {{\v{S}}trumbelj, Erik and Kononenko, Igor},
  title = {An Efficient Explanation of Individual Classifications Using Game Theory},
  journal = {Journal of Machine Learning Research},
  volume = {11},
  pages = {1--18},
  year = {2010},
  url = {https://www.jmlr.org/papers/v11/strumbelj10a.html}
}

@inproceedings{lundberg2017unified,
  author = {Lundberg, Scott M. and Lee, Su-In},
  title = {A Unified Approach to Interpreting Model Predictions},
  booktitle = {Advances in Neural Information Processing Systems},
  volume = {30},
  year = {2017},
  url = {https://proceedings.neurips.cc/paper_files/paper/2017/hash/8a20a8621978632d76c43dfd28b67767-Abstract.html}
}

@inproceedings{covert2020understanding,
  author = {Covert, Ian and Lundberg, Scott and Lee, Su-In},
  title = {Understanding Global Feature Contributions with Additive Importance Measures},
  booktitle = {Advances in Neural Information Processing Systems},
  volume = {33},
  year = {2020},
  url = {https://proceedings.neurips.cc/paper/2020/hash/c7bf0b7c1a86d5eb3be2c722cf2cf746-Abstract.html}
}

@inproceedings{janzing2020feature,
  author = {Janzing, Dominik and Minorics, Lenon and Bl{\"o}baum, Patrick},
  title = {Feature Relevance Quantification in Explainable {AI}: A Causal Problem},
  booktitle = {Proceedings of the 23rd International Conference on Artificial Intelligence and Statistics},
  volume = {108},
  series = {Proceedings of Machine Learning Research},
  pages = {2907--2916},
  year = {2020},
  url = {https://proceedings.mlr.press/v108/janzing20a.html}
}

@inproceedings{covert2021improving,
  author = {Covert, Ian and Lee, Su-In},
  title = {Improving {KernelSHAP}: Practical Shapley Value Estimation Using Linear Regression},
  booktitle = {Proceedings of the 24th International Conference on Artificial Intelligence and Statistics},
  volume = {130},
  series = {Proceedings of Machine Learning Research},
  pages = {3457--3465},
  year = {2021},
  url = {https://proceedings.mlr.press/v130/covert21a.html}
}

@inproceedings{williamson2020efficient,
  author = {Williamson, Brian and Feng, Jean},
  title = {Efficient Nonparametric Statistical Inference on Population Feature Importance Using Shapley Values},
  booktitle = {Proceedings of the 37th International Conference on Machine Learning},
  volume = {119},
  series = {Proceedings of Machine Learning Research},
  pages = {10282--10291},
  year = {2020},
  url = {https://proceedings.mlr.press/v119/williamson20a.html}
}

@article{teneggi2023shapxrt,
  author = {Teneggi, Jacopo and Bharti, Beepul and Romano, Yaniv and Sulam, Jeremias},
  title = {{SHAP-XRT}: The Shapley Value Meets Conditional Independence Testing},
  journal = {Transactions on Machine Learning Research},
  year = {2023},
  url = {https://openreview.net/forum?id=WFtTpQ47A7}
}

@inproceedings{goldwasser2025statistical,
  author = {Goldwasser, Jeremy and Hooker, Giles},
  title = {Statistical Significance of Feature Importance Rankings},
  booktitle = {Proceedings of the 41st Conference on Uncertainty in Artificial Intelligence},
  volume = {286},
  series = {Proceedings of Machine Learning Research},
  pages = {1476--1496},
  year = {2025},
  url = {https://proceedings.mlr.press/v286/goldwasser25a.html}
}

@misc{whitehouse2026statistical,
  author = {Whitehouse, Justin and Sawarni, Ayush and Syrgkanis, Vasilis},
  title = {Statistical Inference and Learning for Shapley Additive Explanations ({SHAP})},
  year = {2026},
  eprint = {2602.10532},
  archivePrefix = {arXiv},
  url = {https://arxiv.org/abs/2602.10532}
}

@misc{kim2025phitest,
  author = {Kim, Dongseok and Choi, Hyoungsun and Rasool, Mohamed Jismy Aashik and Oh, Gisung},
  title = {$\phi$-test: Global Feature Selection and Inference for Shapley Additive Explanations},
  year = {2025},
  eprint = {2512.07578},
  archivePrefix = {arXiv},
  url = {https://arxiv.org/abs/2512.07578}
}

@article{neufeld2022treevalues,
  author = {Neufeld, Anna C. and Gao, Lucy L. and Witten, Daniela M.},
  title = {Tree-Values: Selective Inference for Regression Trees},
  journal = {Journal of Machine Learning Research},
  volume = {23},
  number = {305},
  pages = {1--43},
  year = {2022},
  url = {https://jmlr.org/papers/v23/21-0722.html}
}

@misc{loftus2015groups,
  author = {Loftus, Joshua R. and Taylor, Jonathan E.},
  title = {Selective Inference in Regression Models with Groups of Variables},
  year = {2015},
  eprint = {1511.01478},
  archivePrefix = {arXiv},
  url = {https://arxiv.org/abs/1511.01478}
}

@article{fryer2021shapley,
  author = {Fryer, Daniel and Str{\"u}mke, Inga and Nguyen, Hien D.},
  title = {Shapley Values for Feature Selection: The Good, the Bad, and the Axioms},
  journal = {IEEE Access},
  volume = {9},
  pages = {144352--144360},
  year = {2021},
  doi = {10.1109/ACCESS.2021.3119110}
}

@misc{nash1994abalone,
  author = {Nash, Warwick and Sellers, Tracy and Talbot, Simon and Cawthorn, Andrew and Ford, Wes},
  title = {Abalone},
  howpublished = {UCI Machine Learning Repository},
  year = {1994},
  doi = {10.24432/C55C7W},
  url = {https://doi.org/10.24432/C55C7W}
}

@misc{yeh1998concrete,
  author = {Yeh, I-Cheng},
  title = {Concrete Compressive Strength},
  howpublished = {UCI Machine Learning Repository},
  year = {1998},
  doi = {10.24432/C5PK67},
  url = {https://doi.org/10.24432/C5PK67}
}

@misc{tsanas2012energy,
  author = {Tsanas, Athanasios and Xifara, Angeliki},
  title = {Energy Efficiency},
  howpublished = {UCI Machine Learning Repository},
  year = {2012},
  doi = {10.24432/C51307},
  url = {https://doi.org/10.24432/C51307}
}

@misc{gas2019emission,
  title = {Gas Turbine {CO} and {NOx} Emission Data Set},
  howpublished = {UCI Machine Learning Repository},
  year = {2019},
  doi = {10.24432/C5WC95},
  url = {https://doi.org/10.24432/C5WC95}
}

@misc{yeh2018realestate,
  author = {Yeh, I-Cheng},
  title = {Real Estate Valuation},
  howpublished = {UCI Machine Learning Repository},
  year = {2018},
  doi = {10.24432/C5J30W},
  url = {https://doi.org/10.24432/C5J30W}
}

@misc{cortez2009wine,
  author = {Cortez, Paulo and Cerdeira, Antonio and Almeida, Fernando and Matos, Telmo and Reis, Jos{\'e}},
  title = {Wine Quality},
  howpublished = {UCI Machine Learning Repository},
  year = {2009},
  doi = {10.24432/C56S3T},
  url = {https://doi.org/10.24432/C56S3T}
}

@inproceedings{duy2021parametric,
  title={Parametric Programming Approach for More Powerful and General Lasso Selective Inference},
  author={Nguyen Le Duy, Vo and Takeuchi, Ichiro},
  booktitle={Proceedings of the 24th International Conference on Artificial Intelligence and Statistics},
  series={Proceedings of Machine Learning Research},
  volume={130},
  pages={901--909},
  year={2021},
  publisher={PMLR},
  url={https://proceedings.mlr.press/v130/nguyen-le-duy21a.html}
}

@article{duy2022generalized,
  title={More Powerful Conditional Selective Inference for Generalized Lasso by Parametric Programming},
  author={Nguyen Le Duy, Vo and Takeuchi, Ichiro},
  journal={Journal of Machine Learning Research},
  volume={23},
  number={300},
  pages={1--37},
  year={2022},
  url={https://jmlr.org/papers/v23/21-0494.html}
}

\end{document}